\documentclass[10pt,twocolumn,letterpaper]{article}

\usepackage{iccv}
\usepackage{times}
\usepackage{epsfig}
\usepackage{graphicx}
\usepackage{amsmath}
\usepackage{amssymb}
\usepackage{booktabs}

\usepackage[pagebackref=true,breaklinks=true,letterpaper=true,colorlinks,bookmarks=false]{hyperref}

\iccvfinalcopy 

\begin{document}

\title{Multi-modal Gated Mixture of Local-to-Global Experts for\\ Dynamic Image Fusion}

\author{Yiming Sun, Bing Cao,  Pengfei Zhu,  Qinghua Hu\\
Tianjin University\\
{\tt\small \{sunyiming1895,caobing,zhupengfei,huqinghua\}@tju.edu.cn}}

\maketitle

\begin{abstract}
    Infrared and visible image fusion aims to integrate comprehensive information from multiple sources to achieve superior performances on various practical tasks, such as detection, over that of a single modality. However, most existing methods directly combined the texture details and object contrast of different modalities, ignoring the dynamic changes in reality, which diminishes the visible texture in good lighting conditions and the infrared contrast in low lighting conditions.
    To fill this gap, we propose a dynamic image fusion framework with a multi-modal gated mixture of local-to-global experts, termed MoE-Fusion, to dynamically extract effective and comprehensive information from the respective modalities.
    Our model consists of a Mixture of Local Experts (MoLE) and a Mixture of Global Experts (MoGE) guided by a multi-modal gate.
    The MoLE performs specialized learning of multi-modal local features, prompting the fused images to retain the local information in a sample-adaptive manner, while the MoGE focuses on the global information that complements the fused image with overall texture detail and contrast.
    Extensive experiments show that our MoE-Fusion outperforms state-of-the-art methods in preserving multi-modal image texture and contrast through the local-to-global dynamic learning paradigm, and also achieves superior performance on detection tasks. 
    Our code will be available: \url{https://github.com/SunYM2020/MoE-Fusion}.
\end{abstract}

\section{Introduction}
Infrared and visible image fusion focus on generating appealing and informative fused images that enable superior performance in practical downstream tasks over that of using single modality alone~\cite{Ma2019InfraredAV,Zhang2021ImageFM,Xu2022U2FusionAU}. In recent years, infrared-visible image fusion has been widely used in many applications, such as autonomous vehicles~\cite{hwang2015multispectral} and unmanned aerial vehicles~\cite{Sun2022DroneBasedRC}.
According to the thermal infrared imaging mechanism, infrared images can be adapted to various lighting conditions but has the disadvantage of few texture details~\cite{ma2020ddcgan,zhao2021efficient}.
By contrast, visible images contain rich texture detail information, but cannot provide clear information in low light conditions.
Therefore, how to design advanced fusion methods such that the fused images preserve sufficient texture details and valuable thermal information has attracted a lot of research attention.

\begin{figure}[!t]
	\centering
	\includegraphics[width=1.0\columnwidth]{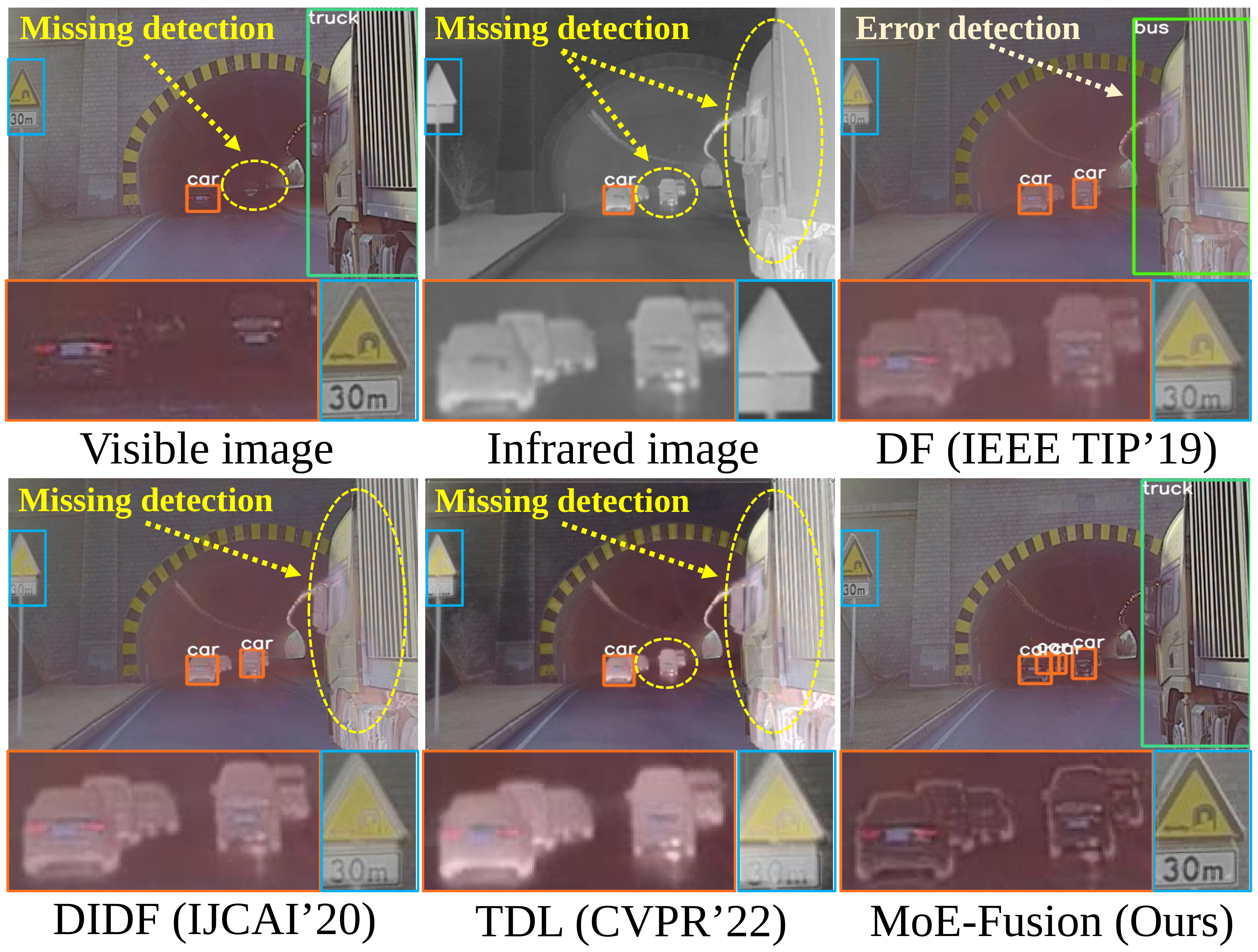}
	\caption{The importance of dynamic image fusion. In the SOTA methods (DF~\cite{li2018densefuse}, DIDF~\cite{ZhaoDIDFuse2020}, and TDL~\cite{liu2022target}), the texture details of objects (\eg, car, truck, and traffic sign) in the fused image are suppressed by the contrast of infrared image, leading to terrible detection results. Benefiting from the dynamic fusion, our method can preserve clear texture details without being interrupted by unsuitable contrast, achieving the best performance.}
	\label{fig1}
\end{figure}

Existing infrared-visible fusion methods~\cite{Xu2020FusionDNAU,Zhang2020RethinkingTI,Wang2022UnsupervisedMI,sun2022detfusion,Xu2022RFNetUN} can be mainly categorized to traditional approaches (image decomposition~\cite{Li2020MDLatLRRAN}, sparse representation~\cite{zhang2018sparse},~\etc) and deep learning-based approaches (autoencoder based methods~\cite{li2018densefuse,ZhaoDIDFuse2020,li2021rfn}, generative adversarial network based approaches~\cite{liu2022target,ma2019fusiongan}, transformer based approaches~\cite{SwinFuse2022,Tang_2022_YDTR},~\etc). 
However, most of these methods directly combined the texture details and object contrast of different modalities, ignoring the dynamic changes in reality, leading to poor fusion results and even weaker downstream task performance than that of a single modality.
As shown in Fig.~\ref{fig1}, the infrared image should adaptively enhance \textit{cars} in dim light while avoiding compromising the textural detail of \textit{truck} in bright light. However, the object textures in these state-of-the-art fusion methods are significantly disturbed by the infrared thermal information due to the lack of dynamic learning of multi-modal local and global information, resulting in terrible object detection performance.

In complex scenes, different modalities have different characteristics: under good lighting conditions, the texture of an object should not be disturbed by thermal infrared information; under low lighting conditions, the contrast of an object also should not be suppressed by the darkness of the visible image. To address this problem, we explore a dynamic fusion framework in a sample-adaptive manner, achieving dynamic learning of multi-modal images from local to global. Fig.~\ref{fig1} shows that the proposed method dynamically balances texture details and contrasts from different sources, and achieves the best detection results. Specifically, we propose a dynamic image fusion framework with a multi-modal gated mixture of local-to-global experts, termed MoE-Fusion, which consists of a Mixture of Local Experts (MoLE) and a Mixture of Global Experts (MoGE) guided by a multi-modal gate. In MoLE, we introduce the attention map generated by an auxiliary network to construct multi-modal local priors and perform dynamic learning of multi-modal local features guided by multi-modal gating, achieving sample-adaptive multi-modal local feature fusion. 
Moreover, MoGE performs dynamic learning of multi-modal global features to achieve a balance of texture details and contrasts globally in the fused images. With the proposed dynamic fusion paradigm from local to global, our model is capable of performing a reliable fusion of different modal images.

The main contributions of this paper are summarized as follows:
\begin{itemize}
\item We propose a dynamic image fusion model, providing a new multi-modal gated mixture of local-to-global experts for reliable infrared and visible image fusion (benefiting from the dynamically integrating effective information from the respective modalities).
\item The proposed model is an effective and robust framework for sample-adaptive infrared-visible fusion from local to global. Further, it prompts the fused images to dynamically balance the texture details and contrasts with the specialized experts.
\item We conduct extensive experiments on multiple infrared-visible datasets, which clearly validate our superiority, quantitatively and qualitatively. Moreover, we also demonstrate our effectiveness in object detection.
\end{itemize}

\section{Related Works}
\label{sec:related}
\subsection{Infrared and Visible Image Fusion}
The infrared and visible image fusion task focuses on generating a fused image containing sufficient information through the learning of multi-modal features~\cite{Ma2019InfraredAV,Zhang2021ImageFM}.
Ma~\etal~\cite{ma2016infrared} defined the goal of image fusion as preserving more intensity information in infrared images as well as gradient information in visible images.
Li~\etal~\cite{li2018densefuse} use the autoencoder to extract multi-modal overall features and fuse them by designed fusion rules, which inspired a series of subsequent works~\cite{Li2020NestFuseAI,li2021rfn}.
Zhao~\etal~\cite{ZhaoDIDFuse2020,zhao2021efficient} proposed the deep learning-based image decomposition methods, which decompose images into background and detail images by high- and low-frequency information, respectively, and then fuse them by designed fusion rules.
Recently, some GAN-based methods~\cite{ma2019fusiongan,ma2020ddcgan,liu2022target} and Transformer-based methods~\cite{SwinFuse2022,Tang_2022_YDTR} have also attracted wide attention.
These works, despite the different approaches adopted, all focus on learning on the representation of the overall multi-modal features. However, they ignore the dynamic changes in reality, which diminishes the visible texture in good lighting conditions and the infrared contrast in low lighting conditions.
We propose a dynamic image fusion framework that enables sample-adaptive fusion from local to global. This approach prompts the fused images to balance the texture details and contrast with specialized experts dynamically.

\begin{figure*}[!t]
	\centering
	\includegraphics[width=2.1\columnwidth]{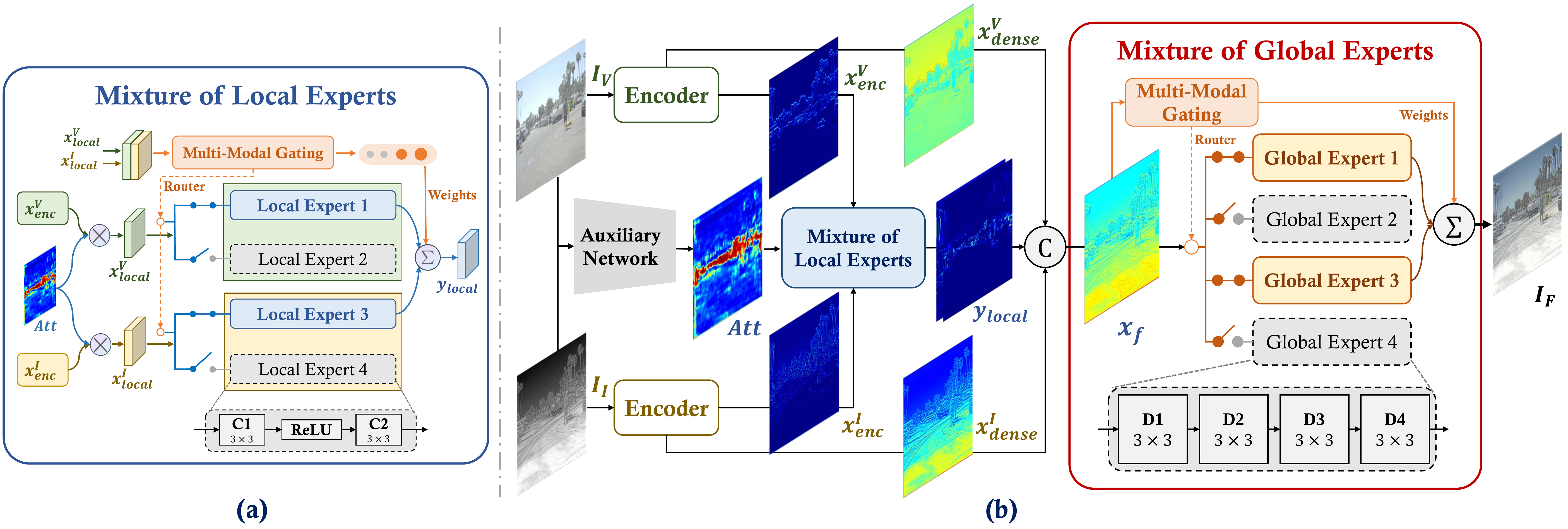}
 \vspace{-5pt}
	\caption{The architecture of MoE-Fusion. MoE-Fusion consists of a Mixture of Local Experts (MoLE), a Mixture of Global Experts (MoGE) guided by the multi-modal gate, a set of infrared and visible feature encoders, and the auxiliary network.}
 \vspace{-12pt}
	\label{fig:MoEFusion}
\end{figure*}

\subsection{Mixture-of-Experts}
Mixture-of-Experts (MoE)~\cite{Jacobs1991AdaptiveMO,Riquelme2021ScalingVW,Mustafa2022MultimodalCL} can dynamically adjust its structure according to different inputs. 
Shazeer~\etal~\cite{Shazeer2017OutrageouslyLN} constructed a sparsely-gated MoE layer that uses a gate network to select multiple experts and assign weights to each selected expert. The final result of the MoE is a weighted sum of the outputs of the different experts. This work also serves as the basis for subsequent research.
Recently, some researchers~\cite{Lepikhin2021GShardSG,Nie2021DensetoSparseGF} have focused on exploring learning mechanisms in MoE, trying to solve the problems of unbalanced expert load and the number of activated experts faced by MoE during training. 
Other researchers~\cite{Fedus2021SwitchTS,Zhu2022UniPerceiverMoELS,Du2022GLaMES,Wang2021VLMoUV} focuses on the combination of MoE and Transformer. They expect to use MoE to build sparse models.
Zhu~\etal~\cite{Zhu2022UniPerceiverMoELS} introduced Conditional MoE into the generalist model and proposed different routing strategies to mitigate the interference between tasks and modalities.
Existing MoE-related methods focus on modeling generic knowledge by using the dynamicity and sparsity of MoE, resulting in each expert does not know what they should be expert in. 
In contrast, we extend the idea of MoE to image fusion tasks for the first time, constructing a multi-modal gated mixture of local-to-global experts, assigning specific tasks to each expert, and enabling sample-adaptive specialized learning, which yields superior performance.

\section{Methods}
\subsection{Overall Architecture}
In this paper, we propose a dynamic image fusion framework with a multi-modal gated mixture of local-to-global experts, termed MoE-Fusion. In Fig.~\ref{fig:MoEFusion}, MoE-Fusion contains two encoders, a Mixture of Local Experts, a Mixture of Global Experts, and the auxiliary network.

In~Fig.~\ref{fig:MoEFusion} (b), we send a pair of infrared image $I_\mathcal{I} \in \mathbb{R}^{H \times W \times 1}$ and visible image $I_\mathcal{V} \in \mathbb{R}^{H \times W \times 3}$ into the infrared and visible encoders ($Enc_\mathcal{I}$ and $Enc_\mathcal{V}$) to extract the features, respectively. 
The structure of encoders follows~\cite{li2018densefuse}. 
The output of the encoder has two parts: the feature maps of the last layer ($x_{enc}^\mathcal{I}$ and $x_{enc}^\mathcal{V}$) and the dense feature maps ($x_{dense}^\mathcal{I}$ and $x_{dense}^\mathcal{V}$). 
More details of the structure are provided in the supplementary material. We send the $x_{enc}^\mathcal{I}$ and $x_{enc}^\mathcal{V}$ to the MoLE along with the attention map which is learning from the auxiliary network. In this paper, we use Faster-RCNN~\cite{ren2015faster} as the auxiliary network.
In MoLE, we send visible and infrared features to specific local experts separately to achieve dynamic integration of local features under the guidance of multi-modal gating.
We concatenate the outputs of MoLE with the dense feature maps as the input to the MoGE. 
Each expert in MoGE has the ability to decode global features, and a multi-modal gate network is used to dynamic select which experts are activated to decode multi-modal fusion features. The final fused image $I_\mathcal{F} \in \mathbb{R}^{H \times W \times 1}$ is generated by a weighted sum mechanism of the different global experts.

The MoE-Fusion is optimized mainly by calculating the pixel loss and gradient loss between the fused image $I_\mathcal{F}$ and two source images ($I_\mathcal{I}$ and $I_\mathcal{V}$). In addition, we also introduce the load loss to motivate each expert to receive roughly equal numbers of training images. The auxiliary detection networks are optimized independently by the detection loss.

\subsection{Mixture of Local Experts}
In infrared-visible image fusion tasks, specialized learning of multi-modal local information by a sample adaptive manner helps to overcome the challenge of multi-modal fusion failure in complex scenes.
To realize this vision, we need to address two questions: (1) How to find local regions in multi-modal images; (2) How to learn the local features dynamically due to the differences in various samples.

As shown in~Fig.~\ref{fig:MoEFusion} (a), we propose a Mixture of Local Experts (MoLE) to dynamically learning the multi-modal local features. We use the auxiliary detection networks with a spatial attention module to learn the attention maps. Then we can find the local regions in multi-modal images according to the guidance of the learned attention maps. Specifically, we introduce attention modules in two auxiliary detection networks for extracting visible attention maps and infrared attention maps, respectively. The modal-specific attention map $Att_\mathcal{V}$/$Att_\mathcal{I}$ is generated by the attention module between the feature extractor and the detection head in the detection networks, which consists of a Conv($1\times1$)-BN-ReLU layer and a Conv($1\times1$)-BN-Sigmoid layer. The $Att_\mathcal{V}$ and $Att_\mathcal{I}$ are concatenated and fed into $2$ convolutional layers, the maximum of which output is the $Att$. In MoLE, we multiply the $x_{enc}^\mathcal{V}$ and $x_{enc}^\mathcal{I}$ with $Att$ to obtain $x_{local}^\mathcal{V}$ and $x_{local}^\mathcal{I}$, respectively. Then we concatenate the $x_{local}^\mathcal{V}$ and $x_{local}^\mathcal{I}$ to get the $x_{local}$, which is the input of the multi-modal gating network. 
The MoLE is consist of a multi-modal gating network $G_{local}$ and a set of $N$ expert networks $\{E_{1}^{local},...,E_{N}^{local}\}$. 
The structure of each expert network is $2$ convolution layers and $1$ ReLU layers.

In MoLE, we flatten the input $x_{local} \in \mathbb{R}^{H \times W \times C}$ to $s_{local} \in \mathbb{R}^{D}$.
The gate network $G_{local}$ takes the vector $s_{local}$ as input and produces the probability of it with respective to $N$ experts. The formalization of the gate network is as follows,
\begin{align}
  G_{local}(s_{local})= softmax(topK(s_{local} \cdot W_{local})),
\end{align}
where $W_{local} \in \mathbb{R}^{D \times N}$ is a learnable weight matrix and the top $K$ outputs are normalized via $softmax$ function. 

To achieve specialized learning of different modalities, we input visible local features $x_{local}^\mathcal{V}$ into one set of expert networks $\{E_{1}^{local},...,E_{N/2}^{local}\}$ and infrared local features $x_{local}^\mathcal{I}$ into another set of non-overlapping expert networks $\{E_{(N/2)+1}^{local},...,E_{N}^{local}\}$.
Each expert network produces its own output $E_{i}^{local}(x_{local}^{j})$.
The final output $y_{local}$ of the MoLE is calculated as follows,
\begin{align}
    y_{local} = \sum_{i=1}^{N}G_{local}(s_{local})_{i}E_{i}^{local}(x_{local}^{j}),
\end{align}
where $j$ represents $\mathcal{I}$ or $\mathcal{V}$. Then we concatenate the $y_{local}$, $x_{dense}^\mathcal{I}$, and $x_{dense}^\mathcal{V}$ to obtain the global multi-modal fusion feature $x_{f}$.

\subsection{Mixture of Global Experts}
Traditional image fusion algorithms use the same network structure and parameters to learn the fusion features of different samples. In contrast, we propose the MoGE to dynamically integrate multi-modal global features, which can adaptively adjust its own structure and parameters when dealing with different samples, thus showing superior advantages in terms of model expressiveness and self-adaptability. The main components of the MoGE include a global multi-modal gating network $G_{global}$ and a set of $N$ expert networks $\{E_{1}^{global},...,E_{N}^{global}\}$. In MoGE, we flatten $x_{f}$ to get $s_{f}$ and feed it into $G_{global}$. The corresponding gating weights of the $N$ expert networks are calculated as follows,
\begin{equation}
    G_{global}(s_{f}) = softmax(topK(s_{f} \cdot W_{global})),
\end{equation}
where $W_{global}$ is a learnable weight matrix and the top $K$ outputs are normalized via $softmax$ distribution.

The structure of each expert network consists of $4$ convolution layers. Each expert takes the global multi-modal fusion feature $x_{f}$ as input to produce its own output $E_{i}^{global}(x_{f})$. The final output $I_\mathcal{F}$ of MoGE is the linearly weighted combination of each expert’s output with the corresponding gating weights. The formalization is as follows,
\begin{equation}
    I_\mathcal{F} = \sum_{i=1}^{N}G_{global}(s_{f})_{i}E_{i}^{global}(x_{f}).
\end{equation}

\subsection{Loss Function}
In MoE-Fusion, we use the fusion loss $\mathcal{L}_{fusion}$ to guide the optimization of the image fusion network, and the auxiliary detection networks are optimized by their respective detection loss~\cite{ren2015faster} ($\mathcal{L}_{det}^\mathcal{V}$ or $\mathcal{L}_{det}^\mathcal{I}$). We end-to-end train the entire framework through these three loss functions. Specifically, the formula of fusion loss is as follows,
\begin{align}
  \mathcal{L}_{fusion}= \mathcal{L}_{pixel}+ \alpha \mathcal{L}_{grad}+ \mathcal{L}_{load},
\end{align}
where the pixel loss $\mathcal{L}_{pixel}$ constrains the fused image to preserve more significant pixel intensities originating from the target images, while the gradient loss $\mathcal{L}_{grad}$ forces the fused image to contain more texture details from different modalities. $\mathcal{L}_{load}$ represents load loss, which encourages experts to receive roughly equal numbers of training examples~\cite{Shazeer2017OutrageouslyLN}. More details about pixel loss, gradient loss and load loss are provided in the supplementary material. $\alpha$ is used to strike a balance between the different loss functions.

\begin{figure}[!t]
	\centering
	\includegraphics[width=1.0\columnwidth]{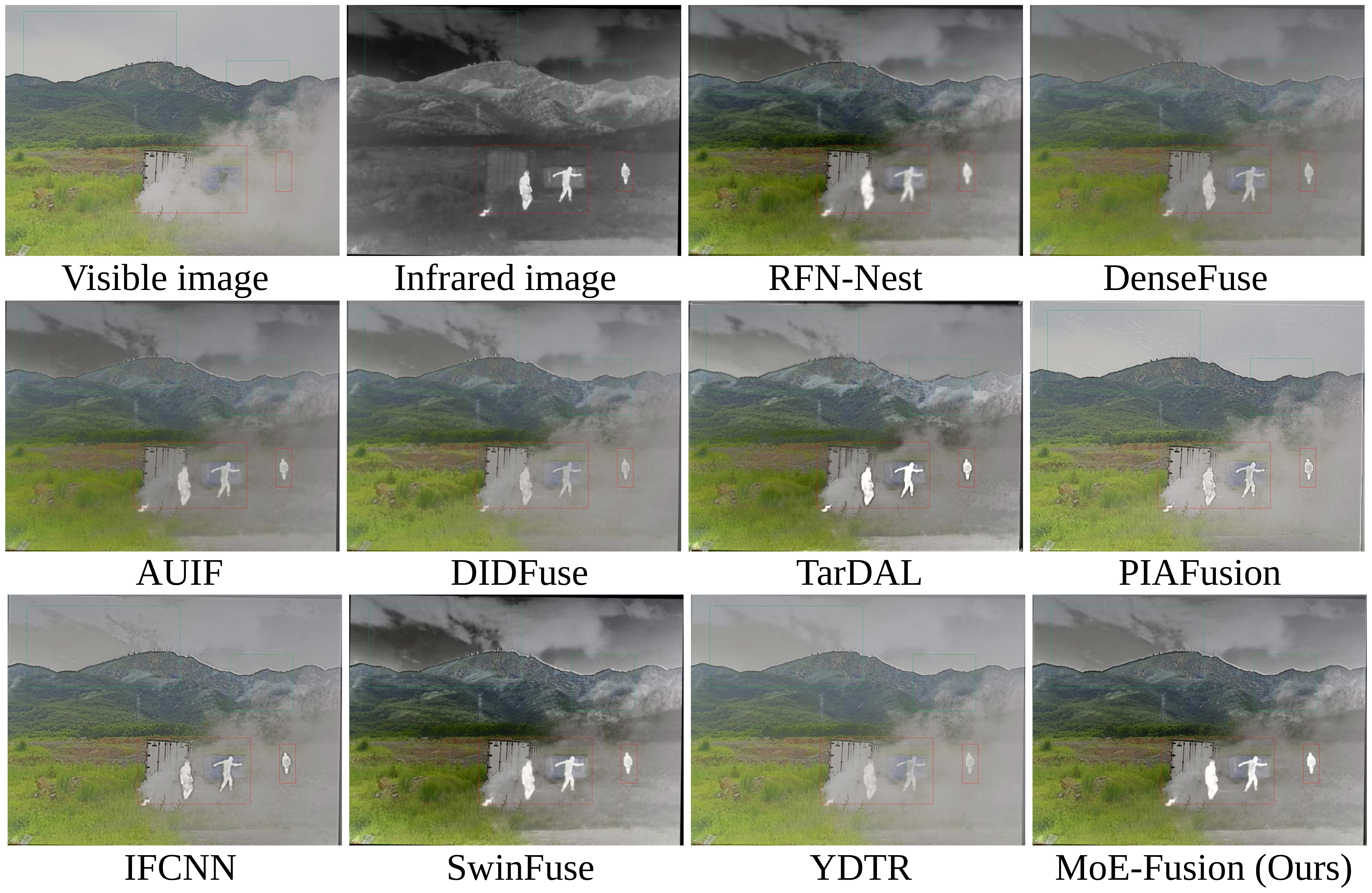}
	\caption{Qualitative comparisons of various methods on representative images selected from the M$^{3}$FD dataset.}
	\label{fig:M3FD_Compared}
\end{figure}

\section{Experiments}
\subsection{Experimental Setting}

\noindent \textbf{Datasets and Partition Protocol.} We conducted experiments on three publicly available datasets: (M$^{3}$FD~\cite{liu2022target}, LLVIP~\cite{jia2021llvip} and FLIR~\cite{fafree}).

{\bf M$^{3}$FD:} It contains $4,200$ infrared-visible image pairs captured by on-board cameras. We used $3,900$ pairs of images for training and the remaining $300$ pairs for evaluation.

{\bf FLIR:} We used the ``aligned" version~\cite{zhang2020multispectral} of FLIR in this work. It contains $5,142$ infrared-visible image pairs captured by on-board cameras. We used $4,129$ image pairs for training and $1,013$ image pairs for evaluation. 

{\bf LLVIP:} The LLVIP dataset contains $15,488$ aligned infrared-visible image pairs, which is captured by the surveillance cameras in different street scenes. We trained the model with $12,025$ image pairs and performed evaluation on $3,463$ image pairs. 

\begin{table}[t]
    \centering
    \caption{Quantitative comparison of our MoE-Fusion with $9$ state-of-the-art methods. Bold \textcolor{red}{red} indicates the best, Bold \textcolor{blue}{blue} indicates the second best, and Bold \textcolor{cyan}{cyan} indicates the third best.}
    \vspace{0.2cm}
    \label{tab:Quantitative}%
    \resizebox{\linewidth}{!}{
        \begin{tabular}{ccccccccc}
            \toprule
            \multicolumn{9}{c}{\textbf{M$^{3}$FD Dataset}~\cite{liu2022target}}        \\
            &EN        &SF         &SD         &MI        &VIF       &AG         &SCD       &$Q_{abf}$       \\ 
            \midrule
            DenseFuse~\cite{li2018densefuse}   &$6.4134$	&$0.0364$	 &$8.5987$	   &$2.9524$	  &$0.6572$	  &$3.0700$	  &$1.5069$	   &$0.3838$\\
            RFN-Nest~\cite{li2021rfn}      &$6.9208$   &$0.0345$	&$9.2984$	&$2.9301$	  &$0.7806$	   &$3.1698$	 &$1.5410$	 &$0.3772$\\
            IFCNN~\cite{zhang2020ifcnn}    &$6.6555$	  &$0.0599$	 &$9.2456$	&$2.9954$	&$0.7522$	&$5.0932$	&\textcolor{blue}{$\mathbf{1.5448}$}	  &$0.5755$ \\
            PIAFusion~\cite{Tang2022PIAFusion}  &$6.8167$	  &\textcolor{blue}{$\mathbf{0.0707}$}	   &\textcolor{blue}{$\mathbf{10.1228}$}   &\textcolor{blue}{$\mathbf{3.8337}$}  &\textcolor{cyan}{$\mathbf{0.8447}$}   &\textcolor{blue}{$\mathbf{5.6560}$}   &$1.3065$    &$0.5540$\\
            DIDFuse~\cite{ZhaoDIDFuse2020}    &$6.6116$	  &$0.0420$	  &$9.3409$	  &$2.9955$    &$0.7382$	&$3.5668$	&\textcolor{red}{$\mathbf{1.5875}$}	   &$0.4342$ \\
            AUIF~\cite{zhao2021efficient}   &$6.5233$	&$0.0399$	&$8.8759$	&$2.9793$	&$0.6796$	&$3.3224$	&$1.5314$	&$0.4124$\\
            SwinFuse~\cite{SwinFuse2022}  &\textcolor{cyan}{$\mathbf{6.9819}$} 	   &\textcolor{cyan}{$\mathbf{0.0696}$}	 &$9.6400$	   &$3.2004$	  &\textcolor{blue}{$\mathbf{0.9114}$}    &\textcolor{cyan}{$\mathbf{5.6234}$}   &$1.5395$   &$0.5166$\\
            YDTR~\cite{Tang_2022_YDTR}   &$6.5397$	 &$0.0496$	  &$9.2631$	&$3.2128$	 &$0.7276$    &$3.8951$   &$1.5076$	   &$0.4812$\\
            TarDAL~\cite{liu2022target}  &\textcolor{red}{$\mathbf{7.1347}$}	 &$0.0528$	&\textcolor{cyan}{$\mathbf{9.6820}$}	&\textcolor{cyan}{$\mathbf{3.2853}$}	&$0.8347$	&$4.1998$	&$1.5334$	 &$0.3858$\\
            \bf{MoE-Fusion}   &\textcolor{blue}{$\mathbf{7.0018}$}	&\textcolor{red}{$\mathbf{0.0715}$}	  &\textcolor{red}{$\mathbf{10.1406}$}	&\textcolor{red}{$\mathbf{4.1949}$}	&\textcolor{red}{$\mathbf{1.0034}$}	  &\textcolor{red}{$\mathbf{5.6742}$}	   &\textcolor{cyan}{$\mathbf{1.5433}$}	&\textcolor{red}{$\mathbf{0.6661}$}\\
            \midrule
            \multicolumn{9}{c}{\textbf{FLIR Dataset}~\cite{zhang2020multispectral}}    \\
            &EN        &SF         &SD         &MI        &VIF       &AG         &SCD       &$Q_{abf}$       \\ 
            \midrule
            DenseFuse~\cite{li2018densefuse}   &$6.9479$	&$0.0304$	 &$10.5928$	   &$2.9254$	  &$0.6413$	  &$3.0524$	  &$1.3132$	   &$0.2947$\\
            RFN-Nest~\cite{li2021rfn}      &\textcolor{cyan}{$\mathbf{7.4277}$}	    &$0.0267$	&\textcolor{blue}{$\mathbf{10.8747}$}	&$2.9669$	  &$0.7260$	   &$2.7664$	 &\textcolor{cyan}{$\mathbf{1.6731}$}	 &$0.2603$\\
            IFCNN~\cite{zhang2020ifcnn}    &$7.1367$    &\textcolor{blue}{$\mathbf{0.0638}$}	&$10.6668$	&$2.9263$	 &$0.7567$	  &\textcolor{red}{$\mathbf{6.1081}$}	 &$1.3521$	&\textcolor{blue}{$\mathbf{0.4894}$}    \\
            PIAFusion~\cite{Tang2022PIAFusion}  &$6.9968$	  &$0.0551$	   &$10.6563$   &\textcolor{blue}{$\mathbf{3.0927}$}  &\textcolor{blue}{$\mathbf{0.8178}$}   &$5.1849$   &$1.1615$    &\textcolor{cyan}{$\mathbf{0.4388}$}\\
            DIDFuse~\cite{ZhaoDIDFuse2020}    &$7.2754$	   &\textcolor{red}{$\mathbf{0.0670}$}	 &\textcolor{red}{$\mathbf{11.6318}$}	&$2.5022$	   &$0.6649$	 &\textcolor{blue}{$\mathbf{5.8815}$}	&$1.4913$	  &$0.3469$  \\
            AUIF~\cite{zhao2021efficient}   &$7.2853$	&$0.0463$	 &$10.2108$	   &$2.8893$	 &$0.7099$	 &$4.4471$	&$1.5823$	 &$0.3240$\\
            SwinFuse~\cite{SwinFuse2022}  &$7.4163$	   &$0.0582$	 &$10.6077$	   &$2.9404$	  &\textcolor{cyan}{$\mathbf{0.8104}$}    &$5.7587$   &\textcolor{blue}{$\mathbf{1.6739}$}  &$0.3751$\\
            YDTR~\cite{Tang_2022_YDTR}   &$6.8948$	 &$0.0364$	  &$10.6571$	&\textcolor{cyan}{$\mathbf{3.0820}$}	 &$0.6749$    &$3.2993$   &$1.3379$	   &$0.3333$\\
            TarDAL~\cite{liu2022target}  &\textcolor{blue}{$\mathbf{7.4866}$}    &$0.0588$    &$10.6948$ &$3.0228$    &$0.7665$    &$5.1955$    &$1.3182$     &$0.3896$\\
            \bf{MoE-Fusion}      &\textcolor{red}{$\mathbf{7.4925}$}	&\textcolor{cyan}{$\mathbf{0.0603}$}	 &\textcolor{cyan}{$\mathbf{10.7007}$}  &\textcolor{red}{$\mathbf{3.1209}$}	&\textcolor{red}{$\mathbf{0.8212}$}	  &\textcolor{cyan}{$\mathbf{5.7604}$}	 &\textcolor{red}{$\mathbf{1.6818}$}	&\textcolor{red}{$\mathbf{0.4991}$}\\
            \midrule
            \multicolumn{9}{c}{\textbf{LLVIP Dataset}~\cite{jia2021llvip}} \\
            &EN        &SF         &SD         &MI        &VIF       &AG         &SCD       &$Q_{abf}$       \\ 
            \midrule
            DenseFuse~\cite{li2018densefuse}   &$6.8314$	&$0.0426$	 &$9.3800$	&$2.6764$	  &$0.6894$	&$3.2640$	 &$1.2109$	 &$0.3093$\\
            RFN-Nest~\cite{li2021rfn}    &$7.1408$	&$0.0300$	 &$9.7184$	&$2.5042$	  &$0.7294$	 &$2.7853$	  &\textcolor{cyan}{$\mathbf{1.4612}$}	 &$0.2287$\\
            IFCNN~\cite{zhang2020ifcnn}      &$7.2139$	  &\textcolor{cyan}{$\mathbf{0.0688}$}	  &\textcolor{cyan}{$\mathbf{9.7633}$}	  &$2.9479$	  &\textcolor{cyan}{$\mathbf{0.7797}$}	&\textcolor{cyan}{$\mathbf{5.4136}$}	  &$1.4269$	  &\textcolor{blue}{$\mathbf{0.5845}$}\\
            PIAFusion~\cite{Tang2022PIAFusion}   &\textcolor{red}{$\mathbf{7.3954}$}	&\textcolor{blue}{$\mathbf{0.0787}$}	&$9.7320$	  &\textcolor{blue}{$\mathbf{3.3690}$}  &\textcolor{blue}{$\mathbf{0.8860}$}	 &\textcolor{blue}{$\mathbf{6.0846}$}  &\textcolor{blue}{$\mathbf{1.5300}$}  &\textcolor{cyan}{$\mathbf{0.5789}$}\\
            DIDFuse~\cite{ZhaoDIDFuse2020}    &$6.0372$  	&$0.0550$	  &$7.8074$ 	&$2.5137$	  &$0.5054$	  &$3.4474$	   &$1.2574$  	&$0.2436$ \\
            AUIF~\cite{zhao2021efficient}      &$6.1947$	&$0.0636$	 &$7.8418$	&$2.3966$	  &$0.5533$	&$3.8588$	  &$1.2840$	   &$0.2764$\\
            SwinFuse~\cite{SwinFuse2022}   &$5.9973$   &$0.0608$  &$7.6525$   &$2.1846$    &$0.5962$   &$3.7344$   &$1.2629$    &$0.2620$\\
            YDTR~\cite{Tang_2022_YDTR}    &$6.6922$	  &$0.0474$	  &$8.8701$	  &$2.9152$	   &$0.6322$  &$3.2043$  &$1.0881$  &$0.2907$\\
            TarDAL~\cite{liu2022target} &\textcolor{cyan}{$\mathbf{7.3504}$}	 &$0.0647$  &\textcolor{blue}{$\mathbf{9.7676}$}	&\textcolor{red}{$\mathbf{3.4655}$}	 &$0.7769$  &$4.6094$   &$1.3607$	 &$0.4431$\\
            \bf{MoE-Fusion}             &\textcolor{blue}{$\mathbf{7.3523}$}	  &\textcolor{red}{$\mathbf{0.0862}$}	&\textcolor{red}{$\mathbf{9.8664}$}	  &\textcolor{cyan}{$\mathbf{3.1061}$}	&\textcolor{red}{$\mathbf{0.9202}$}	   &\textcolor{red}{$\mathbf{6.1316}$} 	&\textcolor{red}{$\mathbf{1.7841}$}	  &\textcolor{red}{$\mathbf{0.5932}$}\\
            \bottomrule
        \end{tabular}}%
\end{table}

\noindent \textbf{Competing methods.}
We compared the $9$ state-of-the-art methods on three publicly available datasets (M$^{3}$FD~\cite{liu2022target}, LLVIP~\cite{jia2021llvip} and FLIR~\cite{fafree}).  In these comparison methods, DenseFuse~\cite{li2018densefuse} and RFN-Nest~\cite{li2021rfn} are the autoencoder-based methods, PIAFusion~\cite{Tang2022PIAFusion} and IFCNN~\cite{zhang2020ifcnn} are the CNN-based methods, TarDAL~\cite{liu2022target} is the GAN-based methods. DIDFuse~\cite{ZhaoDIDFuse2020} and AUIF~\cite{zhao2021efficient} are the deep learning-based image decomposition methods. SwinFuse~\cite{SwinFuse2022} and YDTR~\cite{Tang_2022_YDTR} are the Transformer-based methods.

\noindent \textbf{Implementation Details.}
We performed experiments on a computing platform with two NVIDIA GeForce RTX 3090 GPUs. We used Adam Optimization to update the overall network parameters with the learning rate of $1.0\times 10^{-4}$. The auxiliary network Faster R-CNN~\cite{ren2015faster} is also trained along with the image fusion pipeline. The training epoch is set to $24$ and the batch size is $4$. The tuning parameter $\alpha$ is set to $10$. For MoLE and MoGE, we set the number of experts is $4$, and sparsely activate the top $2$ experts. 

\noindent \textbf{Evaluation Metrics.}
We evaluated the performance of the proposed method based on qualitative and quantitative results. The qualitative evaluation is mainly based on the visual effect of the fused image. A good fused image needs to have complementary information of multi-modal images.
The quantitative evaluation mainly uses quality evaluation metrics to measure the performance of image fusion. We selected $8$ popular metrics, including the entropy (EN)~\cite{Roberts2008AssessmentOI}, spatial frequency (SF)~\cite{Eskicioglu1995ImageQM}, standard deviation (SD), mutual information (MI)~\cite{Qu2002InformationMF}, visual information fidelity (VIF)~\cite{Han2013ANI}, average gradient (AG)~\cite{Cui2015DetailPF}, the sum of the correlations of differences (SCD)~\cite{aslantas2015new}, and gradient-based similarity measurement ($Q_{abf}$)~\cite{Xydeas2000ObjectiveIF}. We also evaluate the performance of the different methods on the typical downstream task, infrared-visible object detection.

\begin{figure}[!t]
	\centering
	\includegraphics[width=1.0\columnwidth]{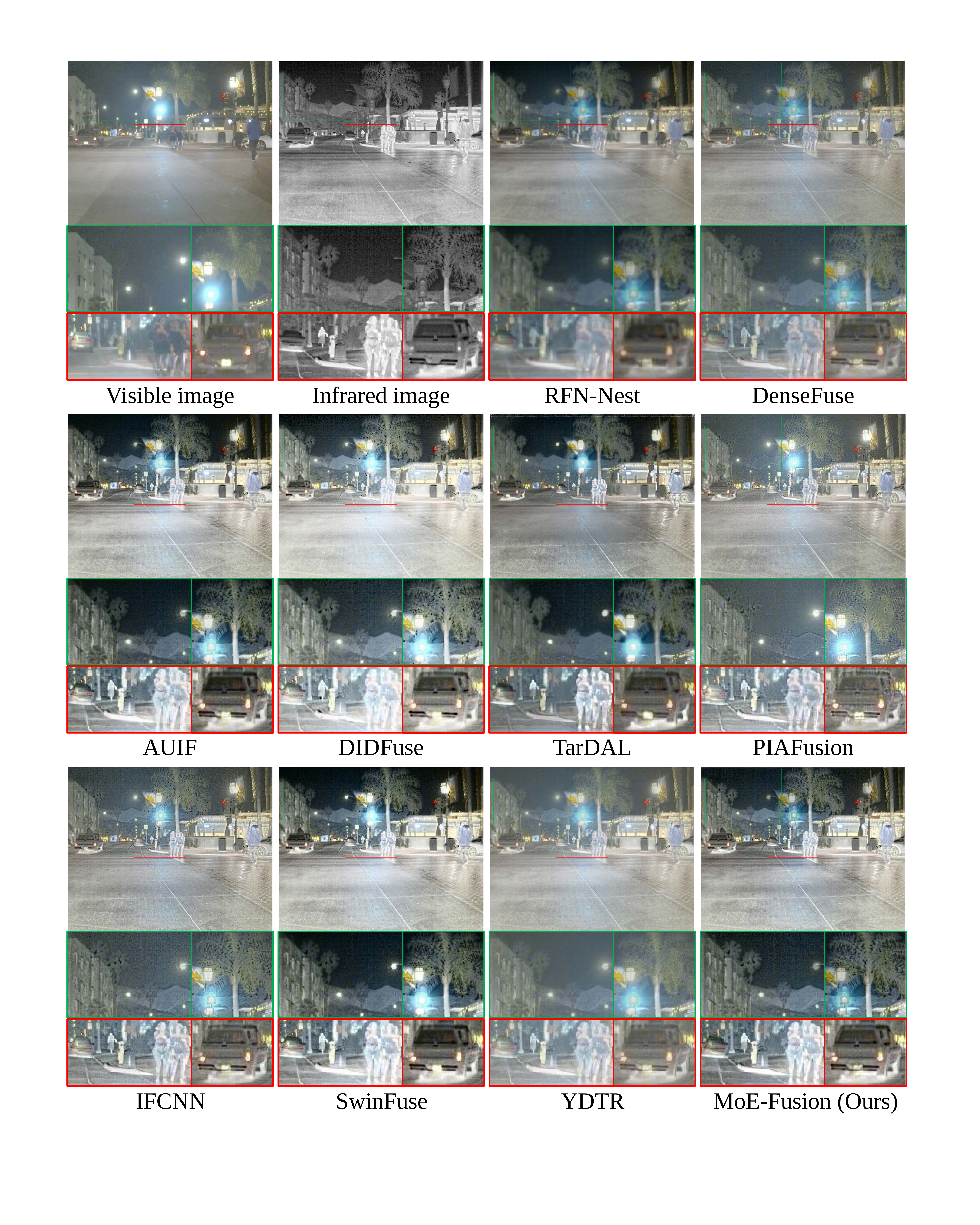}
	\caption{Qualitative comparisons of various methods on representative images selected from the FLIR dataset.}
	\label{fig:FLIR_Compared}
\end{figure}

\subsection{Evaluation on the M$^{3}$FD dataset}
\noindent \textbf{Quantitative Comparisons.}
Table~\ref{tab:Quantitative} presents the results of the quantitative evaluation on the M$^{3}$FD dataset, where our method achieves the best in $7$ metrics and the second and third best performance in the remaining metrics, respectively. In particular, it shows overwhelming advantages on VIF, MI,and $Q_{abf}$, which indicates that our fusion results contain more valuable information and are more beneficial to the visual perception effect of human eyes. The highest SF, SD and AG also indicate that our fusion results preserve sufficient texture detail and contrast. Such superior performance is attributed to the proposed dynamic learning framework from local to global, which achieves state-of-the-art fusion performance through the sample adaptive approach.

\begin{figure}[!t]
	\centering
	\includegraphics[width=1.0\columnwidth]{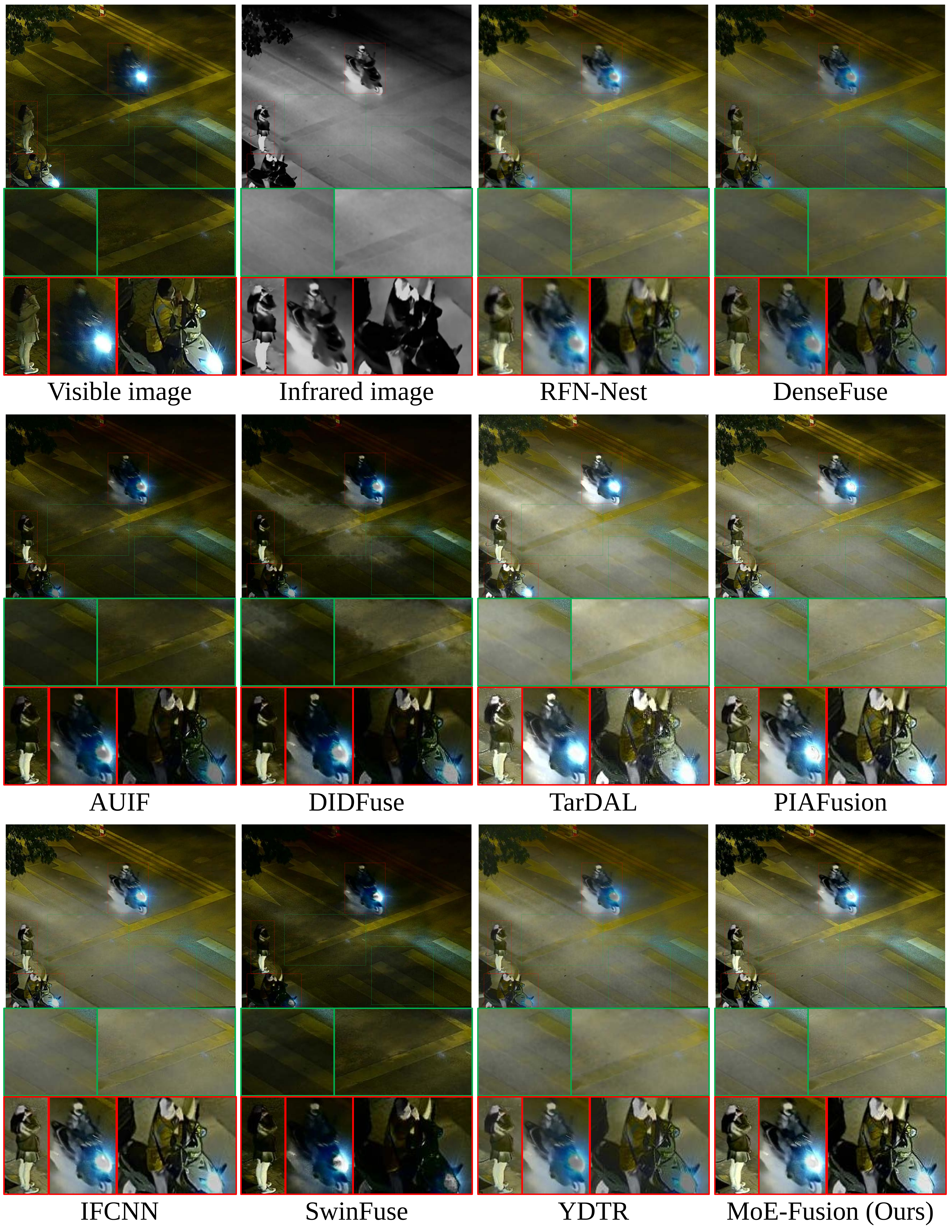}
	\caption{Qualitative comparisons of various methods on representative images selected from the LLVIP dataset.}
	\label{fig:LLVIP_Compared}
\end{figure}

\noindent \textbf{Qualitative Comparisons.}
To better shown the superiority of the proposed method, we assigned the color information of the $3$-channel visible image to the single-channel fused image through the color space conversion between RGB and YCbCr. We mark the background region with the green rectangular box and the foreground region with the red rectangular box. 
As shown in Fig.~\ref{fig:M3FD_Compared}, our fusion results have the best results in both local and global regions. Compared to PIAFusion, YDTR, AUIF, IFCNN and DIDFuse, our fusion results show better contrast, and our fusion results show better texture details compared to TarDAL and SwinFuse. In daytime scenes, our fusion results can be adaptively learned with sufficient texture detail and significant contrast, for example on containers and people. Especially for people, our method effectively avoids the effect of smoke and best preserves the contrast information of infrared, and in local regions, we effectively preserve the rich texture information of containers, outperforming other methods. In global regions such as mountains, grass and sky, our fusion results also effectively retain texture information, which indicates that our method has better visual effects due to effective dynamic learning of local and global information of different modalities.

\subsection{Evaluation on the FLIR dataset}
\noindent \textbf{Quantitative Comparisons.}
Table~\ref{tab:Quantitative} reports the performance of the different methods on the FLIR dataset for $8$ metrics. Our method achieves the best results in $5$ metrics. Among them, the highest EN and MI indicate that our method can preserve abundant information of the multi-modal images well. The best performance of our method on SCD and $Q_{abf}$ also indicates that our fusion results can better learn the multi-modal complementary information and edge information, which makes our fusion results have better foreground-background contrast and richer texture details. Moreover, the highest VIF also demonstrates that our method can generate fused images with better visual effects that are beneficial to human observation. In addition, the third best results on SF, SD, and AG also indicate that our method is highly competitive. The quantitative results on the FLIR dataset also validate the superiority of our method in dynamically fusing multi-modal complementary information from the local to the global.

\noindent \textbf{Qualitative Comparisons.}
To better shown the superiority of the proposed method, we assigned the color information of the $3$-channel visible image to the single-channel fused image through the color space conversion between RGB and YCbCr. 
We mark the background region with the green rectangular box and the foreground region with the red rectangular box. We also show their zoomed-in effects for easier comparison. As shown in Fig.~\ref{fig:FLIR_Compared}, our fusion results have the best results in both local and global regions. In night scenes, our fusion results adaptively learn sufficient texture detail and contrast, for example on buildings, trees, mountains and traffic lights. Especially for the traffic lights, our method effectively avoids the effects of glare and best preserves the entire outline of the traffic lights. On the local regions, our fusion results preserve the best contrast information of the pedestrians and the rich detail information of the vehicles. These comparisons illustrate that our method has better visual effects due to the effective dynamic learning of local information from different modalities. The superiority of the proposed MoE-Fusion also reveals that specialized knowledge of multi-modal local and global in fusion networks can effectively improve fusion performance.

\subsection{Evaluation on the LLVIP dataset}
\noindent \textbf{Quantitative Comparisons.}
The quantitative results of the different methods on the LLVIP dataset are reported in~Table~\ref{tab:Quantitative}. Our method outperforms all the compared methods on $6$ metrics and achieved the second and third best results on the remaining $2$ metrics, respectively. Specifically, the highest SF and AG we achieved indicate that the proposed method preserves richer texture details in the multi-modal images. As well as the highest SD also indicates that our fusion results can contain the highest contrast information between the foreground and the background. SCD and $Q_{abf}$ denote the complementary information and edge information transferred from multi-modal images to fused image, respectively, and our highest results on these two metrics indicate that our method can learn more valuable information from multi-modal images. Moreover, the highest VIF also means that our method can generate the most appealing fused images that are more suitable for human vision. These quantitative results demonstrate that the proposed MoE-Fusion can effectively learn multi-modal knowledge and generate informative and appealing fusion results.

\noindent \textbf{Qualitative Comparisons.}
To better shown the superiority of the proposed method, we assigned the color information of the $3$-channel visible image to the single-channel fused image through the color space conversion between RGB and YCbCr. 
We mark the background region with the green rectangular box and the foreground region with the red rectangular box. We also show their zoomed-in effects for easier comparison.
As shown in Fig.~\ref{fig:LLVIP_Compared}, we can find that the proposed method best preserves the texture details of the local and global in the multi-modal image compared with the state-of-the-art methods, while highlighting the contrast information of the local dynamically. Specifically, for the background region, our fusion results show the sharpest effect on the edge texture of the zebra crossing. For the foreground region, our fusion results preserve the most significant contrast and the richest texture detail in pedestrians and cyclists. Qualitative comparisons show that our MoE-Fusion can balance the texture details and contrasts with the specialized experts dynamically. 

\begin{table}[t]
    \centering
    \caption{Ablation studies on three infrared-visible datasets.}
    \vspace{0.2cm}
    \label{tab:ablation experiment}%
    \resizebox{\linewidth}{!}{
        \begin{tabular}{ccccccccc}
            \toprule
            \multicolumn{9}{c}{\textbf{M$^{3}$FD Dataset}~\cite{liu2022target}} \\
            &EN        &SF         &SD         &MI        &VIF       &AG         &SCD       &$Q_{abf}$       \\ 
            \midrule
            w/o MoLE    &$6.7856$	&$0.0692$	  &$9.1636$	  &$2.7214$	   &$0.8100$	 &$5.5509$	  &$1.5153$	 &$0.6535$\\
            w/o MoGE     &$6.8351$	  &$0.0695$   &$9.2491$   &$2.8138$	   &$0.8261$	 &$5.6521$    &$1.5346$     &$0.6375$\\
            Att-Local       &$6.8656$	  &$0.0693$	  &$9.1894$   &$2.7320$	  &$0.8271$	  &$5.5809$	  &$1.5190$	   &$0.6390$\\
            \bf{MoE-Fusion}      &$\mathbf{7.0018}$	&$\mathbf{0.0715}$ &$\mathbf{10.1406}$   &$\mathbf{4.1949}$	   &$\mathbf{1.0034}$	  &$\mathbf{5.6742}$	 &$\mathbf{1.5433}$	&$\mathbf{0.6661}$\\
            \midrule
            \multicolumn{9}{c}{\textbf{FLIR Dataset}~\cite{zhang2020multispectral}}        \\
            &EN        &SF         &SD         &MI        &VIF       &AG         &SCD       &$Q_{abf}$       \\ 
            \midrule
            w/o MoLE    &$6.9077$    &$0.0661$   &$9.8090$	&$2.5039$	  &$0.5812$	    &$4.9358$	  &$1.1739$	 &$0.4732$\\
            w/o MoGE     &$7.2740$	  &$0.0847$	  &$9.6034$	  &$2.7067$	  &$0.7621$	   &$5.5960$	&$1.6260$	 &$0.5110$\\
            Att-Local        &$7.2528$	  &$0.0858$	  &$9.5483$	  &$2.7129$	  &$0.8599$	   &$6.1099$	&$1.6393$	 &$0.5735$\\
            \bf{MoE-Fusion}             &$\mathbf{7.3523}$  &$\mathbf{0.0862}$	&$\mathbf{9.8664}$  &$\mathbf{3.1061}$	&$\mathbf{0.9202}$	   &$\mathbf{6.1316}$ 	&$\mathbf{1.7841}$	  &$\mathbf{0.5932}$\\
            \midrule
            \multicolumn{9}{c}{\textbf{LLVIP Dataset}~\cite{jia2021llvip}} \\
            &EN        &SF         &SD         &MI        &VIF       &AG         &SCD       &$Q_{abf}$       \\ 
            \midrule
            w/o MoLE    &$6.9021$	&$0.0587$	  &$10.2859$	 &$2.3666$	 &$0.5371$	 &$5.6961$	  &$1.0976$	&$0.3187$\\
            w/o MoGE     &$7.3172$	  &$0.0548$   &$10.6654$   &$2.9208$	&$0.7747$	  &$5.2525$    &$1.6629$     &$0.4796$\\
            Att-Local       &$7.3070$	  &$0.0602$	  &$10.6403$   &$2.6673$	 &$0.7448$	  &$5.7265$	&$1.4788$	   &$0.4838$\\
            \bf{MoE-Fusion}      &$\mathbf{7.4925}$	&$\mathbf{0.0603}$ &$\mathbf{10.7007}$   &$\mathbf{3.1209}$	&$\mathbf{0.8212}$	  &$\mathbf{5.7604}$	 &$\mathbf{1.6818}$	&$\mathbf{0.4991}$\\
            \bottomrule
    \end{tabular}}%
\end{table}

\subsection{Ablation Study}
We conducted ablation studies on the M$^{3}$FD, LLVIP, and FLIR datasets and reported the results in Table~\ref{tab:ablation experiment}.

\noindent\textbf{MoLE.}
To verify the effectiveness of MoLE, we removed MoLE from MoE-Fusion and then extracted multi-modal features by two encoders only, and they were sent to MoGE after concatenation. As shown in Table~\ref{tab:ablation experiment}, all the metrics show a significant decrease after removing MoLE, indicating that MoLE is very effective in learning multi-modal texture details and contrast information. Among others, the decrease on SCD also shows it is difficult to learn complementary multi-modal local information sufficiently without MoLE, which strongly supports our motivation to design MoLE. These results also demonstrate that specialized learning of multi-modal local features can be beneficial to improve image fusion performance.

\noindent\textbf{MoGE.}
We replaced the MoGE with a common decoder with the same structure as that of a single expert in the MoGE. As shown in Table~\ref{tab:ablation experiment}, all the metrics appear to be significantly decreased, strongly verifying that the MoGE can help fused images preserve more contrast and texture detail information. Moreover, these results also demonstrate that MoGE can effectively motivate the image fusion network to dynamically adapt to different samples to learn better feature representations and thus achieve better fusion performance.

\noindent\textbf{Attention-based Local Feature.}
We want to explore how the fusion performance changes when dynamic learning is not performed for local features and only local feature priors constructed by attention maps are used as local features. 
We designed an attention-based local feature learning module (Att-Local), and to keep the output channels consistent, we followed the Att-Local module with a $1 \times 1$ convolution layers. In Table~\ref{tab:ablation experiment}, all metrics cannot exceed the results of MoE-Fusion with the use of Att-Local, but most of them are higher than {\it w/o MoLE}, which demonstrates on the one hand that our proposed MoLE is indeed effective, and on the other hand that the dynamic integration of the effective information from the respective modalities is beneficial to improve the performance of multi-modal image fusion.

\begin{figure}[!t]
	\centering
	\includegraphics[width=1.0\columnwidth]{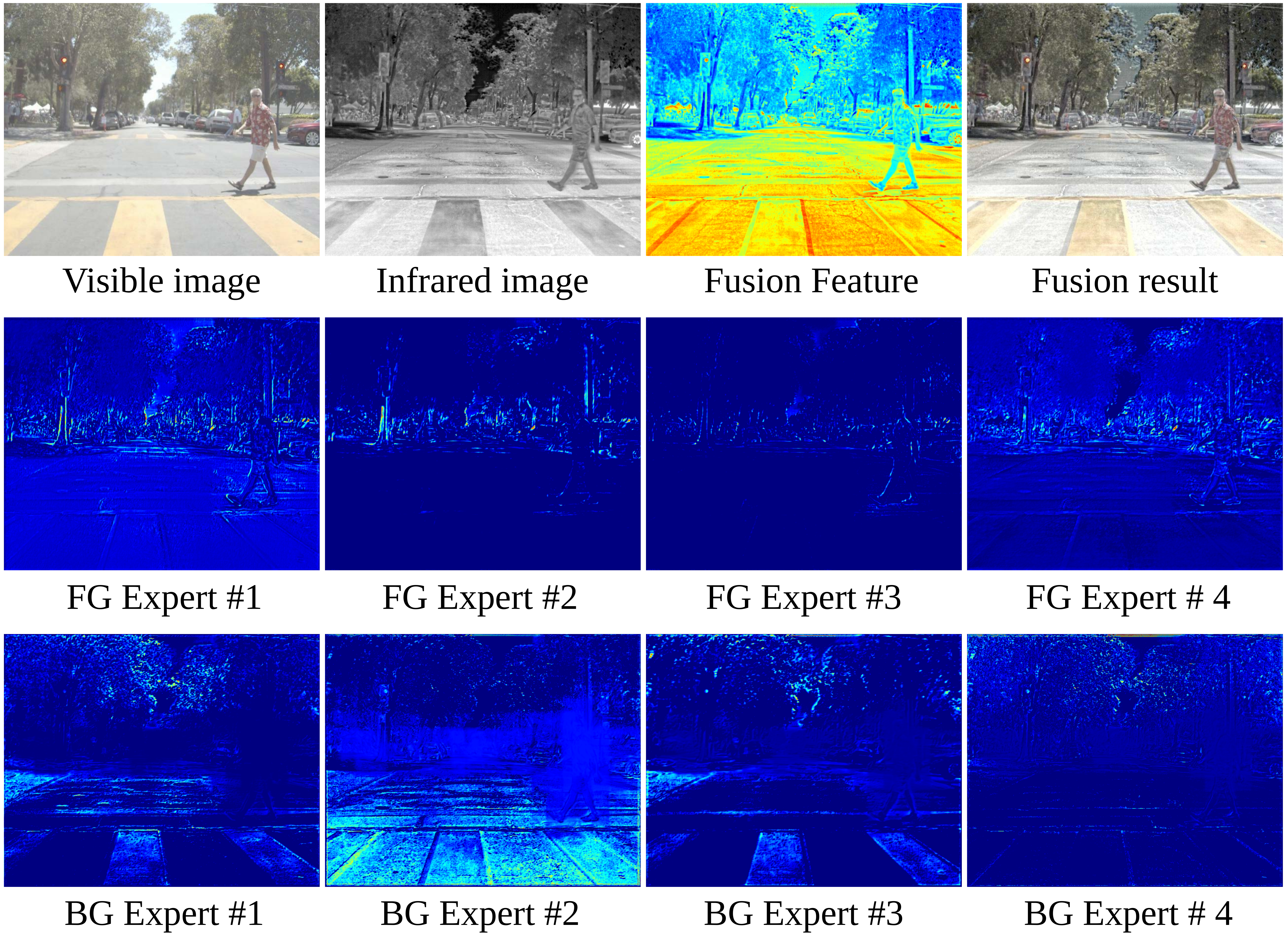}
	\caption{Visualization of the features learned by each expert in MoLE. FG denotes foreground, BG denotes background.}
	\label{fig:MoE_Vis}
\end{figure}

\subsection{Analysis and Discussion}
\noindent\textbf{Visualization of MoLE.}
In the MoLE, we can obtain local feature priors according to the attention map $Att$, which we define as foreground local features (FG), and we also use $1-Att$ to obtain background local features (BG). We visualize what each expert has learned in MoLE. As shown in~Fig.~\ref{fig:MoE_Vis}, we see that the four foreground local experts can clearly learn foreground information, while the four background local experts can also learn rich background features. These results show that MoLE can successfully let each expert know what it should specialize in.

\noindent\textbf{Detection Evaluation.}
Good fused images should have better performance in downstream tasks. For different image fusion methods, we perform the evaluation on the object detection task and use the mean average precision (mAP)~\cite{everingham2015pascal} as the metric.
Following~\cite{9623476}, we first train the object detection model using infrared images and visible images, and then we input the fused images generated by different image fusion methods into the object detection model for inference and evaluate their detection performance. In this paper, we use Faster R-CNN~\cite{ren2015faster} as the object detection algorithm and set the IoU (Intersection over Union) threshold for evaluation to $0.5$. According to Table~\ref{tab:Detection-FLIR}, our MoE-Fusion outperforms all the compared methods and achieves the highest mAP. It is worth noting that our method has an overwhelming advantage on all categories, which demonstrate the proposed dynamic image fusion method is more beneficial for downstream tasks. The detection evaluation on other datasets is provided in the supplementary material.

\begin{figure}[!t]
	\centering
	\includegraphics[width=1.0\columnwidth]{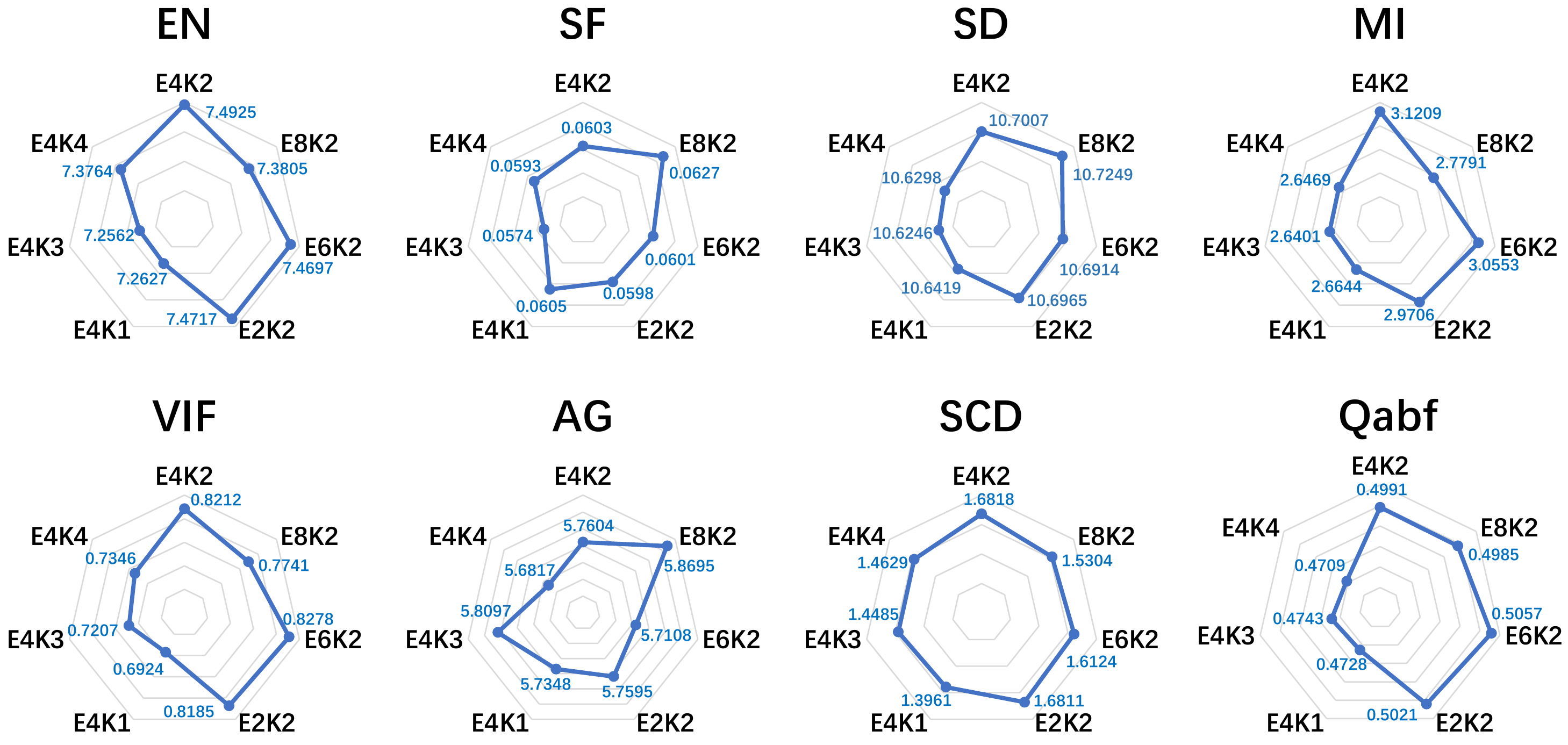}
	\caption{Experiments on the number of experts.}
	\label{fig:Expert_number}
\end{figure}

\begin{table}
  \centering
  \caption{Object detection evaluation on the FLIR dataset. \textcolor{red}{Red} indicates the best.}
  \vspace{0.2cm}
  \label{tab:Detection-FLIR}
  \small
  \renewcommand
  \tabcolsep{6pt}
  \begin{tabular}{ccccc}
    \toprule
    Methods   &car  &person  &bicycle   &mAP\\
    \midrule
    Visible    &$55.90$  &$27.20$  &$33.20$  &$38.70$ \\
    Infrared   &$68.70$  &$45.60$  &$30.80$  &$48.40$ \\
    DenseFuse~\cite{li2018densefuse}   &$65.80$   &$44.70$   &$31.40$   &$47.30$\\
    RFN-Nest~\cite{li2021rfn}          &$58.80$   &$36.50$  &$26.20$  &$40.50$  \\
    IFCNN~\cite{zhang2020ifcnn}        &$67.80$  &$45.70$   &$34.70$   &$49.40$\\
    PIAFusion~\cite{Tang2022PIAFusion}        &$62.60$   &$40.90$   &$31.50$ &$45.00$\\
    DIDFuse~\cite{ZhaoDIDFuse2020}     &$64.20$     &$41.70$    &$32.10$   &$46.00$\\
    AUIF~\cite{zhao2021efficient}      &$61.70$   &$36.80$   &$29.60$   &$42.70$ \\
    SwinFuse~\cite{SwinFuse2022}   &$65.60$	   &$41.30$	   &$29.70$	   &$45.60$\\
    YDTR~\cite{Tang_2022_YDTR}   &$65.00$	 &$44.20$	  &$32.50$  &$47.20$\\
    TarDAL~\cite{liu2022target}   &$62.50$	 &$35.40$	  &$30.80$   &$42.90$\\
    \bf{MoE-Fusion}     &\textcolor{red}{$\mathbf{72.90}$}  &\textcolor{red}{$\mathbf{55.00}$}  &\textcolor{red}{$\mathbf{39.50}$}  &\textcolor{red}{$\mathbf{55.80}$} \\
  \bottomrule
\end{tabular}
\end{table}

\noindent\textbf{Number of Experts.}
As shown in~Fig.~\ref{fig:Expert_number}, we performed $4$ sets of experiments on the FLIR dataset, E2k2, E4K2, E6K2, and E8K2, to explore the effect of the number of experts on the fusion results. As an example, E4K2 means that the MoE contains $4$ experts and sparsely selects top $2$ experts for integration. We found that E4K2 was higher than E2K2 in $7$ metrics, E6K2 in $6$ metrics, and E8K2 in $5$ metrics, suggesting that a higher number of experts may not be better. In addition, we also set up $3$ sets of experiments, E4K1, E4K3 and E4K4, to verify the effect of sparse selection of experts in MoE on the fusion results. In~Fig.~\ref{fig:Expert_number}, we find that E4K2 can outperform E4K1 and E4K3 in $7$ metrics and E4K4 in all metrics. Therefore, in this work, we set $4$ experts for each MoE and sparsely select $2$ experts for integration.

\begin{figure}[!t]
	\centering
	\includegraphics[width=1.0\columnwidth]{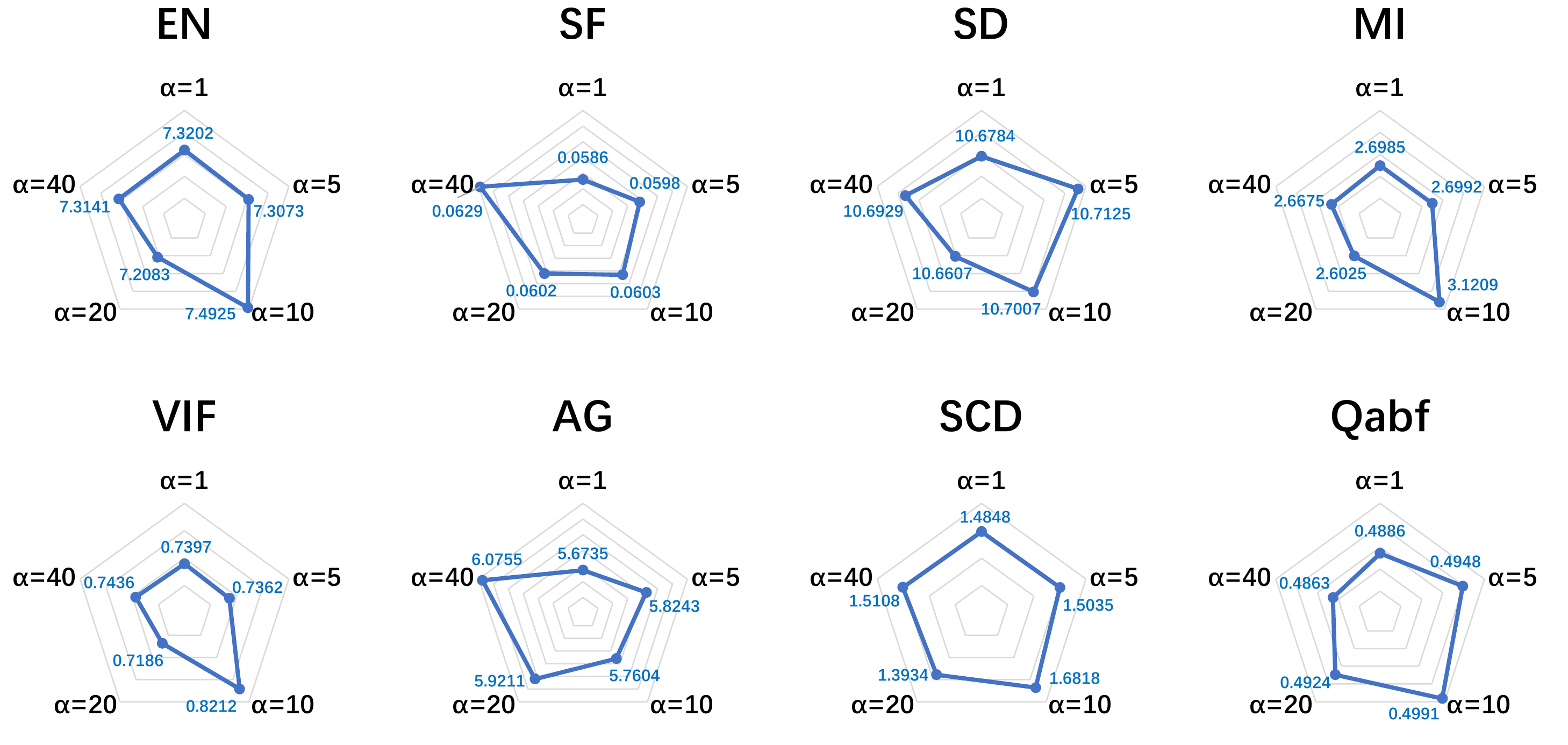}
	\caption{Hyperparameter sensitivity analysis on the FLIR dataset. }
	\label{fig:Hyperparameters}
\end{figure}

\noindent\textbf{Hyperparameter.}
We conducted the hyperparameter experiments on the FLIR dataset to explore the effect of tuning parameter $\alpha$. As shown in~Fig.~\ref{fig:Hyperparameters}, we choose five values for $\alpha$ ($1$, $5$, $10$, $20$, and $40$) and experiment with them in turn. 
When $\alpha$ is less than $10$, the fusion results fail to exceed the performance on all metrics with $\alpha$ equal to $10$.
When $\alpha$ is greater than $10$, there is an improvement in only $2$ metrics (SF and AG) compared to $\alpha$ equal to $10$, but the other $6$ metrics show a decrease. Therefore, in this work, we set $\alpha$ to $10$ to obtain better results.

\section{Conclusion}
In this paper, we propose a novel dynamic image fusion framework with a multi-modal gated mixture of local-to-global experts (MoE-Fusion), which can produce reliable infrared-visible image fusion results. Our framework focuses on dynamically integrating effective information from different source modalities by performing sample-adaptive infrared-visible fusion from local to global. The MoE-Fusion model dynamically balances the texture details and contrasts with the specialized local experts and global experts. The experimental results on three challenging datasets demonstrate that the proposed MoE-Fusion outperforms state-of-the-art methods in terms of visual effects and quantitative metrics. Moreover, we also validate the superiority of our MoE-Fusion in the object detection task. In future work, we will explore leveraging the uncertainty of different images to guide the fusion, and investigate developing an uncertainty-gated MoE paradigm for dynamic image fusion.

\appendix

\begin{figure*}[!t]
	\centering
	\includegraphics[width=2.0\columnwidth]{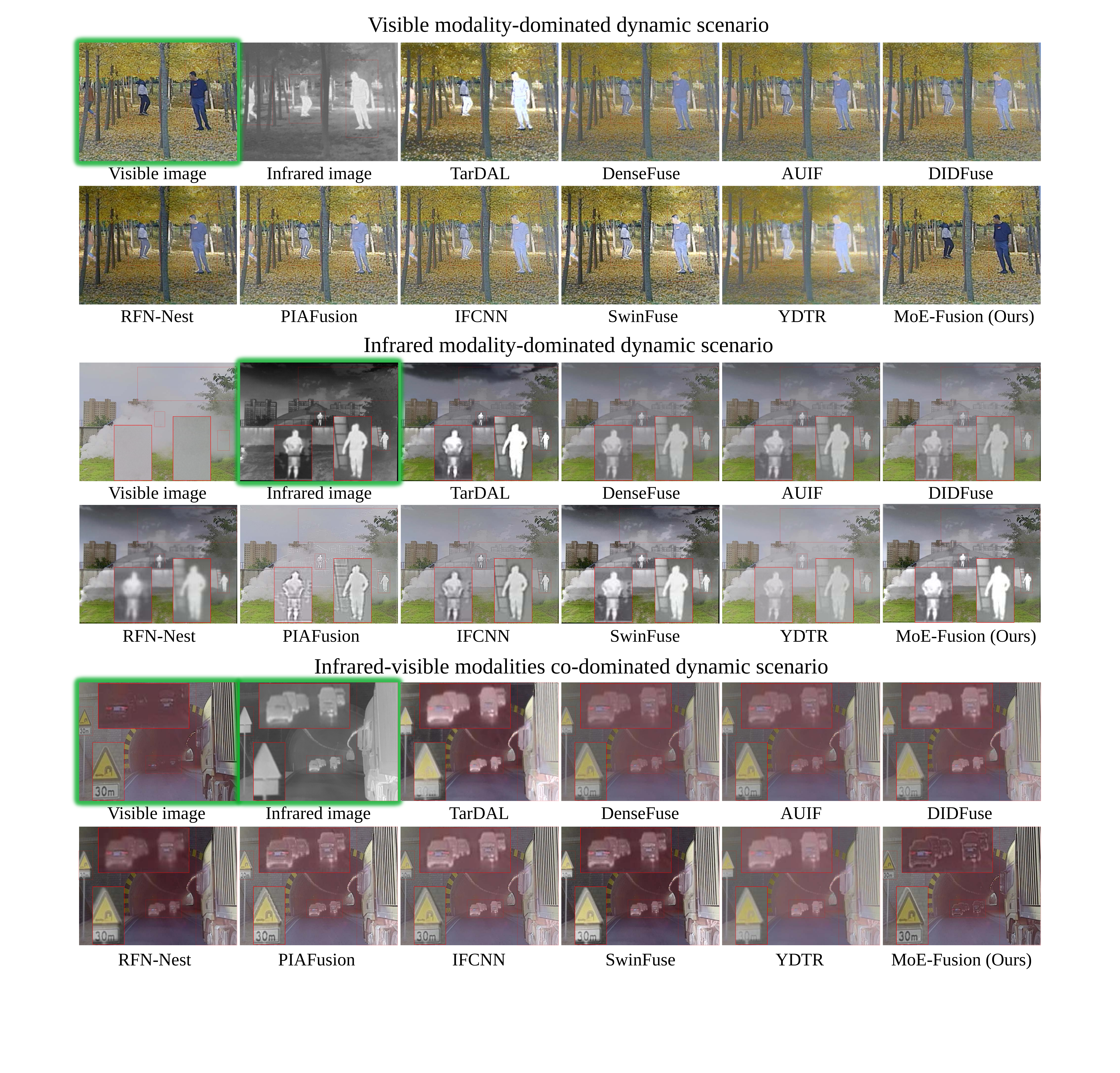}
	\caption{Qualitative comparison of various methods in three representative dynamic scenarios. We use the green bounding box to mark the corresponding modality as dominant.}
	\label{fig:dynamicfusion}
\end{figure*}

\section{Supplemental Materials}
In this supplementary material, we first discuss the results of dynamic fusion in three representative dynamic scenarios, then explain the justification for dynamic fusion from a gradient perspective, and finally, we show that the proposed dynamic fusion model MoE-Fusion has a significant contribution to downstream tasks through a qualitative evaluation on an object detection task. In addition, we perform a qualitative analysis of the ablation studies to strongly confirm the effectiveness of the proposed components. Moreover, we show the evaluation results of different methods on the LLVIP and M$^{3}$FD datasets for the object detection task. To show the performance advantages of the proposed method in local regions, we also perform a quantitative comparison of local regions (foreground and background) of the fused images on three datasets. The above supplementary contents fully reveal that the proposed MoE-Fusion enables sample-adaptive fusion and achieves the most significant advantages on downstream tasks by relying on the powerful dynamic learning capability. At the end of the supplemental material, we also provide details of the two encoders in the proposed MoE-Fusion, details of the fusion loss for optimizing the proposed fusion network, and details of the color space conversion. Our Code will be released for reliably reproducing.

\begin{figure}[!t]
	\centering
	\includegraphics[width=1.0\columnwidth]{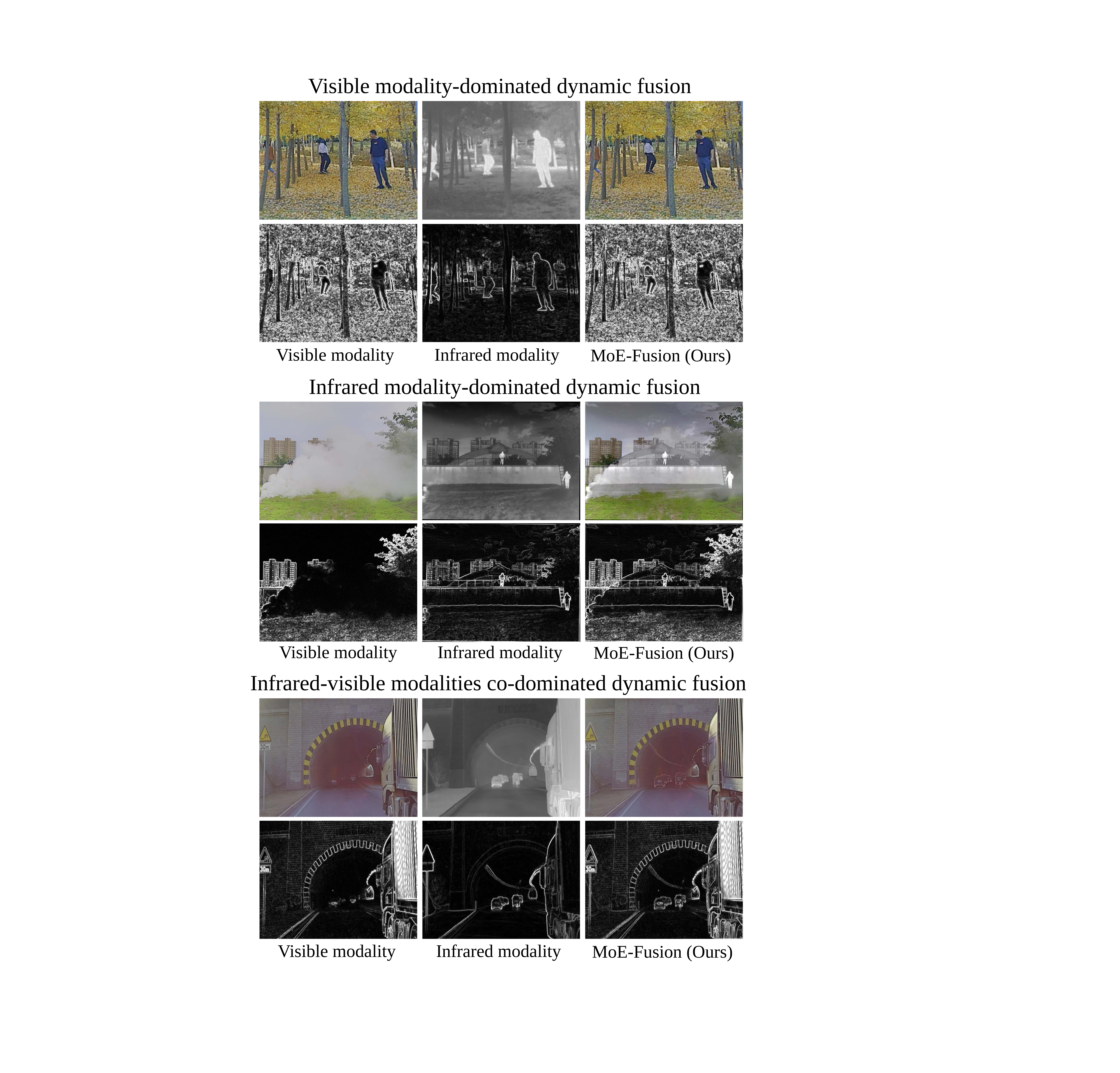}
	\caption{Analysis of dynamic fusion from a gradient perspective. We use the {\it Sobel} operator to generate the gradient maps of the visible modality, the infrared modality, and the fused results, respectively.}
	\label{fig:gradient_Performance}
        \vspace{-0.3cm}
\end{figure}

\begin{figure*}[!t]
	\centering
	\includegraphics[width=2.0\columnwidth]{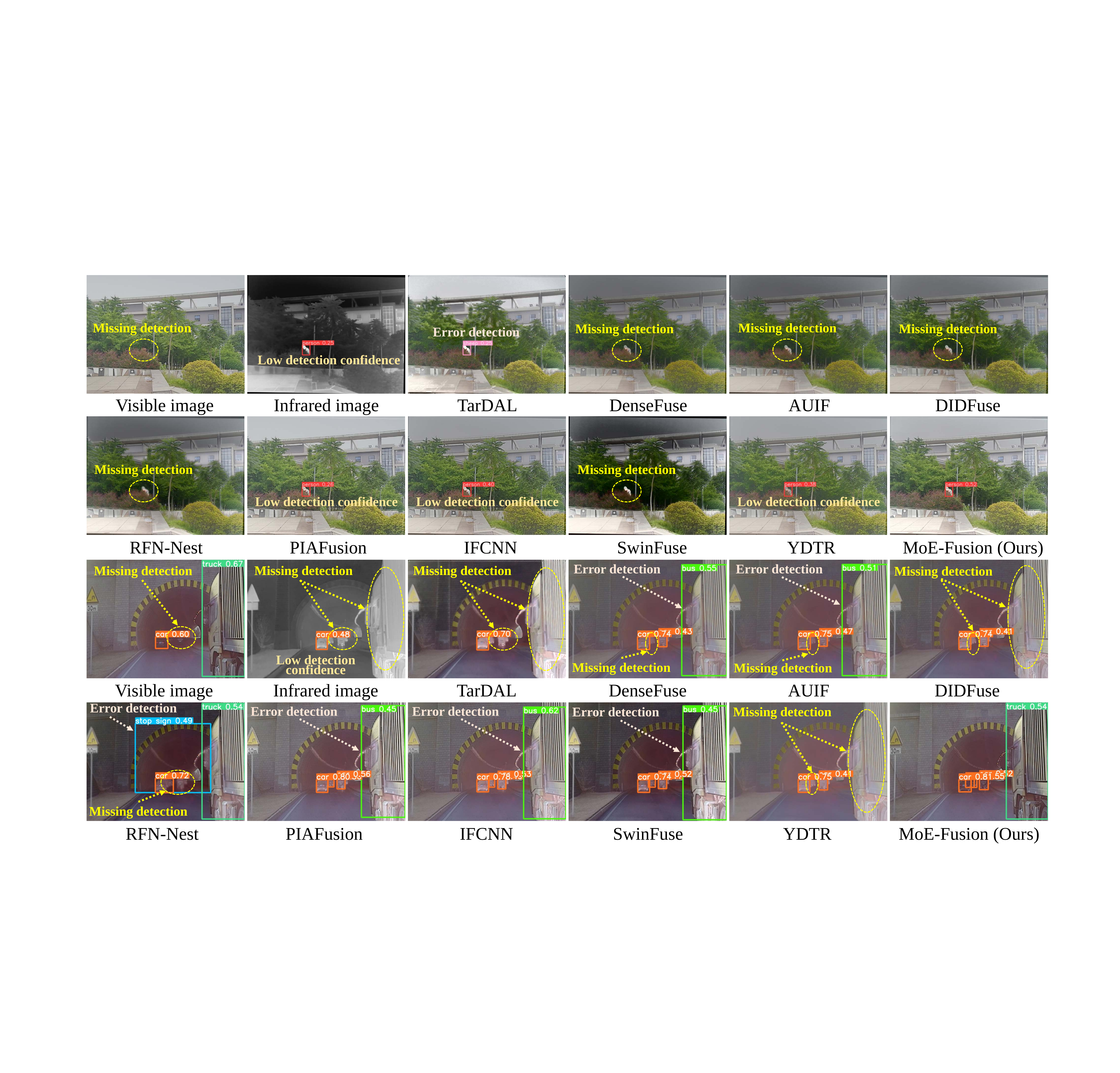}
	\caption{Qualitative comparison of various methods on detection performance.}
	\label{fig:Detection_Performance}
        \vspace{-0.3cm}
\end{figure*}

\section{Discussion on Dynamic Fusion}
\subsection{Qualitative Comparison in Complex Scenarios}
In complex scenes, different modalities have different characteristics: the texture of an object should not be disturbed by thermal infrared information when it is clearly visible in the visible image; the contrast of the object should not be suppressed by the unfavorable information of the visible image (smoke, darkness,~\etc) when the visible image is low quality. Therefore, we qualitatively compare different methods on three representative scenes (visible modality-dominated, infrared modality-dominated, and visible-infrared modalities co-dominated).

We mark the focused regions using red rectangular boxes. As shown in Fig.~\ref{fig:dynamicfusion}, in the visible modality-dominated scenario, our model dynamically learns effective textural details in infrared and visible images, avoiding the redundant infrared contrasts affecting the visible textures. However, the competing methods TarDAL, PIAFusion, IFCNN, SwinFuse, YDTR, AUIF, and DIDFuse failed to dynamically learn the effective information from two different modalities. In daytime scenes, the competing method TarDAL suffered from the over-contrast on the object, which seriously affected the local textures. As a comparison, our fusion results can be adaptively learned with sufficient texture detail and reliable contrast for people. 

In the infrared modality-dominated scenario, our model dynamically learns the significant contrast of the object, avoiding the suppression of the thermal information in the infrared image by the smoke in the visible image. Unfortunately, competing methods DenseFuse, AUIF, DIDFuse, RFN-Nest, PIAFusion, IFCNN, SwinFuse, and YDTR suffer from varying degrees of suppression of object contrast from smoke due to indiscriminately and directly fusing information from multiple modalities. Although the competing method TarDAL preserves contrast due to enhanced learning of objects, black shadows, and noise appear in background areas such as the sky and tree branches, which interfere with the overall texture detail. Our approach achieves a sample-adaptive dynamic fusion from local to global, effectively preserving the high-value information of different modalities.

In the visible-infrared modalities co-dominated scenario, our model dynamically preserves the texture details of traffic signs and trucks while effectively learning the contrast information of vehicles in the tunnel. The competing methods TarDAL, YDTR, DenseFuse, AUIF, PIAFusion, and DIDFuse incorrectly fuse thermal information of traffic signs and trucks from infrared images into the final results, leading to serious over-exposure problems. The remaining methods have similar issues, resulting in fusion results with varying degrees of distortion on the trucks and traffic signs. The experimental results demonstrate our effectiveness in dynamically fusing multi-modal knowledge. This is particularly valuable for potential downstream applications, such as image fusion-based object detection. 

\subsection{Analysis from the Gradient Perspective}
In three representative dynamic scenarios, we use the {\it Sobel} operator to extract the gradient information for the visible modality, the infrared modality, and our fusion results, respectively.  As shown in Fig.~\ref{fig:gradient_Performance}, we can find that in scenes dominated by visible modality, the visible modality has the richest texture details compared to the infrared modality, so our method tends to mainly learn information from the visible modality. In scenes dominated by infrared modality, the infrared modality can also provide rich high-frequency information because it is not affected by smoke occlusion, so our method takes the infrared modality as the main learning reference to make up for the shortcomings of the visible modality. In scenes where infrared-visible modalities co-dominate, trucks and traffic signs have rich high-frequency gradient information in the visible modality, while the gradient information of vehicles in the tunnel needs to be reflected in the infrared modality. Our fusion results dynamically learn the complementary information of gradients in the two modalities, preserving the complete gradient information for trucks, traffic signs, vehicles, ~\etc. According to the analysis of gradient perspective, the amount of effective information contained in different modalities is promising to be the basis of model dynamic learning, and in future work, we will take this as a clue to explore the trustworthy dynamic fusion of multi-modal images.

\subsection{Qualitative Evaluation on the Object Detection}
Following~\cite{liu2022target}, we utilize YOLOv5 as the detection model.
As shown in Fig.~\ref{fig:Detection_Performance}, our model dynamically learns the knowledge of multi-modal images in a sample-adaptive manner and achieves the best detection performance in the two example scenes. Competing methods directly combined the texture details and object contrast of different modalities, ignoring the dynamic changes in reality. 
For example, in the first scenario, the competing method TarDAL over-learned the thermal information of infrared, making the detection model incorrectly detect a person as a sheep. The rest of the competing methods do not make full use of the valid information of the different modalities, leading to the problem that the detection model cannot detect the object or has low detection confidence (below $0.5$). In contrast, our method has the highest detection confidence, outperforming the unimodal detection results.
In the second detection scenario, most competing methods also suffer from infrared modalities that do not adequately preserve the texture details of the trucks and cars, leading to missed and error detection problems. Our method dynamically learns key information of different modalities from local to global by the sample-adaptive manner to generate fused images that are more suitable for downstream tasks.

\begin{figure}[!t]
	\centering
	\includegraphics[width=1.0\columnwidth]{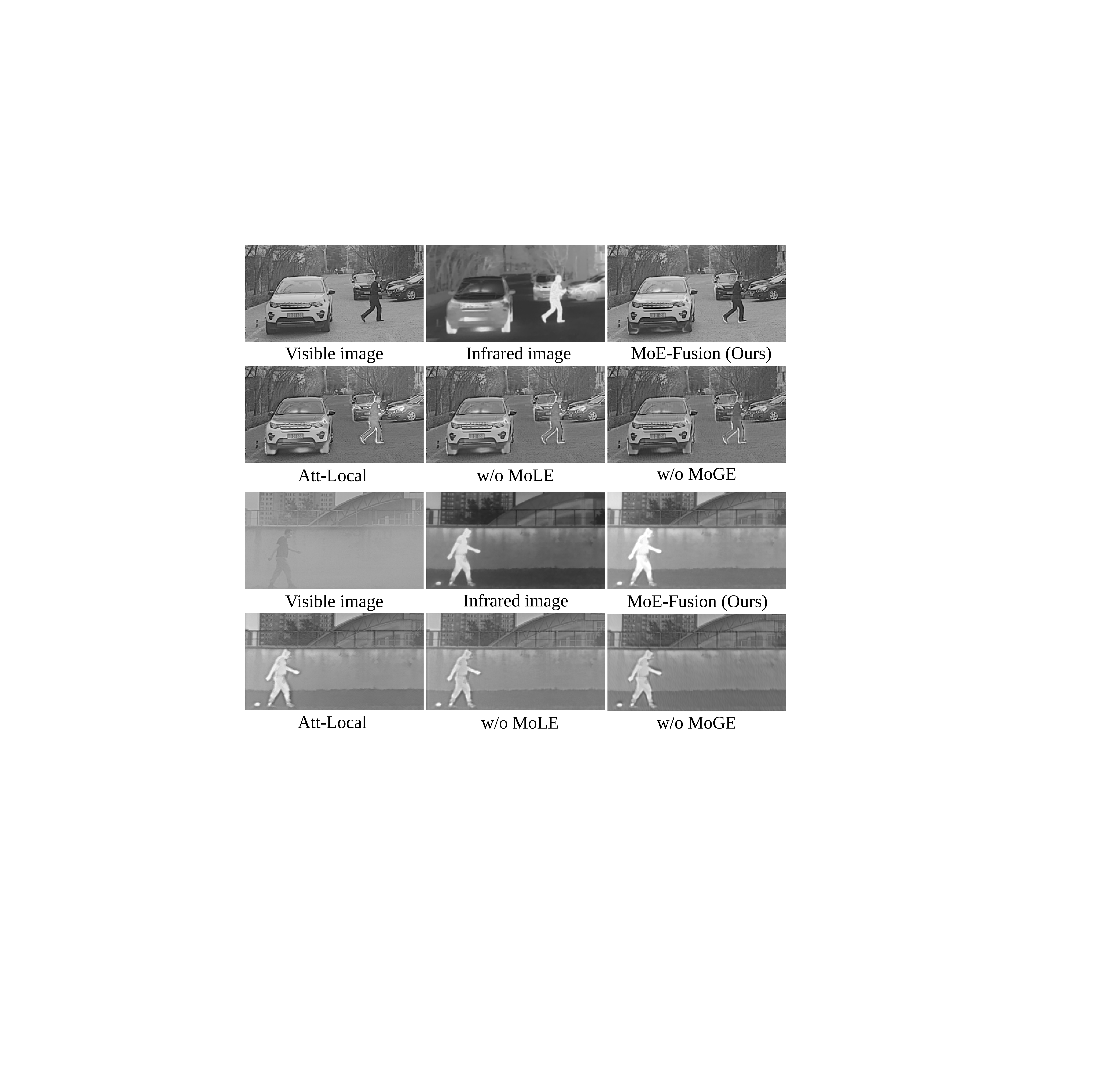}
	\caption{Qualitative analysis of ablation studies for the proposed MoE-Fusion on M$^{3}$FD dataset.}
	\label{fig:ablabtion_Performance}
        \vspace{-0.3cm}
\end{figure}

\subsection{Qualitative Analysis of Ablation Studies}
As shown in Fig.~\ref{fig:ablabtion_Performance}, we qualitatively analyze the results of the ablation studies on the M$^{3}$FD dataset. We use grayscale images to show the two scenarios of visible modality dominance and infrared modality dominance.
In visible modality-dominated scenes, the fusion results of MoE-Fusion can effectively preserve the rich texture details of the visible modality and avoid the effects of over-exposure of the infrared modality on pedestrians. Removing MoLE from MoE-Fusion clearly shows that the lack of dynamic learning of multi-modal local information causes the over-exposure problem of pedestrians. 
Replacing MoLE with Att-Local also leads to bad results, which strongly demonstrates the importance of MoLE to dynamically learn local information in multi-modal images.
In addition, removing MoGE in MoE-Fusion also causes the fusion results to be disturbed by the infrared information of the object due to the lack of global dynamic learning, which also proves the effectiveness of MoGE. 
In infrared modality-dominated scenes, the fusion results of MoE-Fusion also preserve the most significant contrast information, and removing either MoLE or MoGE leads to a loss of contrast information, making the model vulnerable to suppression by smoke in visible images. Replacing MoLE in MoE-Fusion with Att-Local also weakens the contrast of pedestrians but is better than w/o MoLE, which also proves the effectiveness of the proposed MoLE.
With the combination of MoLE and MoGE, MoE-Fusion can dynamically balance texture details and contrast from local to global with specialized experts, which achieves superior fusion performance.

\begin{table}
  \centering
  \caption{Object detection evaluation on the M$^{3}$FD dataset. \textcolor{red}{Red} indicates the best. ``Motor" is short for Motorcycle.}
  \vspace{0.2cm}
  \label{tab:Detection-M3FD}
  \small
  \renewcommand
  \tabcolsep{2pt}
  \begin{tabular}{cccccccc}
    \toprule
    Methods   &People  &Car  &Bus  &Motor  &Lamp  & Truck   &mAP\\
    \midrule
    Visible    &$17.9$  &$70.7$  &$78.1$  &$34.8$  &$33.2$   &$49.9$   &$47.4$\\
    Infrared   &$59.2$  &$64.1$  &$54.5$  &$19.3$  &$18.3$  &$42.7$  &$43.0$\\
    DenseFuse~\cite{li2018densefuse}     &$43.4$    &$73.0$   &$76.7$   &$38.7$    &$21.0$   &$45.6$   &$49.7$\\
    RFN-Nest~\cite{li2021rfn}            &$31.8$    &$68.4$   &$74.5$   &$31.1$    &$19.8$   &$41.4$   &$44.5$\\
    IFCNN~\cite{zhang2020ifcnn}          &$49.2$    &$74.6$   &$77.4$   &$40.0$    &$21.2$   &$48.3$   &$51.8$\\
    PIAFusion~\cite{Tang2022PIAFusion}   &$49.6$    &$73.5$   &$73.2$   &$41.0$    &$29.5$   &$51.0$   &$52.9$\\
    DIDFuse~\cite{ZhaoDIDFuse2020}       &$44.4$    &$72.7$   &$76.2$   &$34.3$    &$23.3$   &$44.7$   &$49.3$\\
    AUIF~\cite{zhao2021efficient}        &$46.8$    &$74.4$   &$78.7$   &$40.0$    &$26.1$   &$47.9$   &$52.3$\\
    SwinFuse~\cite{SwinFuse2022}   &$52.3$	   &$75.5$	   &$79.3$	   &$42.2$    &$29.9$   &$47.1$   &$54.4$\\
    YDTR~\cite{Tang_2022_YDTR}     &$44.5$	 &$74.0$	  &$76.2$  &$33.6$    &$22.0$   &$52.3$   &$50.4$\\
    TarDAL~\cite{liu2022target}    &$59.1$	 &$66.7$	  &$67.2$   &$30.8$    &$12.5$   &$42.3$   &$46.4$\\
    \bf{MoE-Fusion}     &\textcolor{red}{$\mathbf{59.7}$}  &\textcolor{red}{$\mathbf{77.1}$}  &\textcolor{red}{$\mathbf{80.2}$}  &\textcolor{red}{$\mathbf{42.7}$}   &\textcolor{red}{$\mathbf{34.1}$}   &\textcolor{red}{$\mathbf{56.1}$}   &\textcolor{red}{$\mathbf{58.3}$}\\
  \bottomrule
\end{tabular}
\end{table}

\begin{table*}
  \caption{Object detection evaluation on the LLVIP dataset. \textcolor{red}{Red} indicates the best. LLVIP only contains the category ``person", so only mAP is shown.}
  \vspace{0.2cm}
  \label{tab:Detection-LLVIP}
  \renewcommand\tabcolsep{9pt}
  \begin{tabular}{ccccccc}
    \toprule
    Methods   &Visible  &Infrared   &DenseFuse~\cite{li2018densefuse}   &RFN-Nest~\cite{li2021rfn}    &IFCNN~\cite{zhang2020ifcnn}   &PIAFusion~\cite{Tang2022PIAFusion}\\
    \midrule
    mAP   &$51.8$  &$85.6$  &$66.2$  &$67.5$  &$85.9$  &$87.3$ \\
    \midrule
    Methods   &DIDFuse~\cite{ZhaoDIDFuse2020}  &AUIF~\cite{zhao2021efficient}   &SwinFuse~\cite{SwinFuse2022}  &YDTR~\cite{Tang_2022_YDTR}  &TarDAL~\cite{liu2022target}  &\bf{MoE-Fusion}\\
    \midrule
    mAP   &$69.9$  &$64.1$  &$53.7$  &$77.4$   &$85.0$  &\textcolor{red}{$\mathbf{91.0}$}\\
  \bottomrule
\end{tabular}
\end{table*}

\section{Detection Evaluation}
According to Table~\ref{tab:Detection-M3FD}, the proposed MoE-Fusion achieves the most superior performance on the M$^{3}$FD dataset, $3.9\%$ ahead of the next best method SwinFuse on mAP, and more than $10\%$ ahead of the results for the infrared/visible modalities. Our method also achieves the best performance in each category. In particular, our performance on Lamp ($34.1\%$) is substantially ahead of all competing methods and infrared modalities. These illustrate that MoE-Fusion fully exploits the power of multi-modal fusion by dynamically learning valid information from multi-modal images.
According to Table~\ref{tab:Detection-LLVIP}, our MoE-Fusion outperforms all the competing methods and achieves the highest mAP. We achieved a $3.7\%$ advantage over the second-best method PIAFusion. Considering that LLVIP is a night scene dataset and thus the detection performance of visible modality is poor, by dynamically fusing infrared and visible images, the upper limit of infrared modality can be broken and achieved beyond the detection performance of single modality.
The results of these detection evaluations demonstrate the proposed dynamic image fusion method is more beneficial for downstream tasks. 

\section{Local Quantitative Comparisons}
To show the advantages of the proposed method in local regions of fused images, we also perform the local quantitative comparisons of the fused images on three datasets. We use the foreground and background areas to represent the local regions separately. Firstly, we use the annotated bounding boxes in the dataset to generate foreground and background masks. Then, we use these masks to extract the foreground and background images of the fusion results separately. In the foreground image, the region except the foreground objects is masked, and their pixel values are set to $0$. In the background image, all the foreground objects are masked, and their pixel values are set to $0$. On three datasets, we quantitatively evaluate the foreground and background images of the different methods separately to compare the fusion performance of the different methods in local regions intuitively. The results are reported in~Table~\ref{tab:FG_SOTA} and Table~\ref{tab:BG_SOTA}, respectively.

\subsection{Local Comparison on the M$^{3}$FD Dataset}
According to Table~\ref{tab:FG_SOTA}, our method achieves the best results on $7$ metrics. In particular, it shows overwhelming advantages on VIF, MI, and $Q_{abf}$, 
which indicates that the fusion results of the proposed method in the local foreground region are more favorable to the visual perception effect of human eyes and also contain more valuable information. The highest EN, SF, SD, and AG also show that our method effectively preserves the texture details and contrast in the local foreground region of the multi-modal image. The experimental results reveal the effectiveness of the proposed MoE-Fusion to dynamically learn the local information of multi-modal images in a sample-adaptive manner.

In Table~\ref{tab:BG_SOTA}, our method achieves the best results on six metrics, the second best on SCD and the third best on EN. Our method dynamically learns the effective information of multi-modal images from local to global and thus can preserve the best high-frequency texture information on the local background region, as demonstrated by the results on SF, SD, and AG. The significant advantages on VIF, MI, and $Q_{abf}$ also show that the proposed method can retain more valuable local background information in multi-modal images while producing fused images that are more suitable for human visual perception. Such superior performance is attributed to the proposed dynamic learning framework from local to global, which achieves state-of-the-art fusion performance through the sample adaptive approach. The excellent performance in the local region demonstrates that MoE-Fusion can balance global and local learning, which will benefit the performance of downstream tasks.

\subsection{Local Comparison on the FLIR Dataset}
In Table~\ref{tab:FG_SOTA}, our method outperforms all the compared methods on $6$ metrics and achieved the third best results on the remaining $2$ metrics, respectively. Specifically, the highest EN, MI, and SD indicate that the proposed method preserves the richest information and the highest contrast on the foreground image. The best results on SCD and $Q_{abf}$ show that our method can prompt the foreground of the fused image to learn the most valuable complementary and edge information from the multi-modal images. In addition, the highest VIF also demonstrates the advantages of the foreground images generated by our method on visual effects. These results effectively demonstrate that the proposed MoE-Fusion has significant advantages in the local foreground quality of fused images due to the specialized learning of multi-modal local information. 

As shown in Table~\ref{tab:BG_SOTA}, our method achieves superiority on $5$ metrics, where the highest EN, MI, SF, and AG represent that our method preserves the richest texture details and pixel intensity information on the local background region of fused images. As well as the highest VIF indicates that our fusion results are also best suited for human vision on the local background region. The results on $Q_{abf}$ and SCD also illustrate that our method is competitive in preserving multi-modal image complementary and edge information. These results also demonstrate that our fusion results are significantly superior to other methods on local background quality due to dynamic specialized learning of multi-modal local information.

\begin{table*}[t]
    \centering
    \caption{Local Quantitative comparison of our MoE-Fusion with $9$ state-of-the-art methods for {\bf the foreground regions of fused images}. Bold \textcolor{red}{red} indicates the best, Bold \textcolor{blue}{blue} indicates the second best, and Bold \textcolor{cyan}{cyan} indicates the third best.}
    \vspace{0.2cm}
    \label{tab:FG_SOTA}%
    \resizebox{\linewidth}{!}{
        \begin{tabular}{ccccccccc}
            \toprule
            \multicolumn{9}{c}{\textbf{M$^{3}$FD Dataset}~\cite{liu2022target}}        \\
            &EN        &SF         &SD         &MI        &VIF       &AG         &SCD       &$Q_{abf}$       \\
            \midrule
            DenseFuse~\cite{li2018densefuse}   &$0.7273$	&$0.0274$	 &$3.5816$	   &$0.7874$	  &$1.1452$	  &$0.5994$	  &$0.4655$	   &$0.5420$\\
            RFN-Nest~\cite{li2021rfn}      &$0.7469$   &$0.0265$	&$3.5744$	&$0.7865$	  &$1.1289$	   &$0.5973$	 &$0.5111$	 &$0.5090$\\
            IFCNN~\cite{zhang2020ifcnn}    &$0.7544$	  &$0.0349$	 &$3.5772$	&$0.7995$	&$1.1732$	 &$0.8649$	&$0.8726$	  &\textcolor{blue}{$\mathbf{0.6492}$} \\
            PIAFusion~\cite{Tang2022PIAFusion}  &\textcolor{blue}{$\mathbf{0.7650}$}	  &\textcolor{blue}{$\mathbf{0.0373}$}	   &$3.5740$   &\textcolor{blue}{$\mathbf{0.8208}$}  &\textcolor{cyan}{$\mathbf{1.1845}$}   &\textcolor{blue}{$\mathbf{0.9019}$}   &\textcolor{red}{$\mathbf{1.0557}$}    &\textcolor{cyan}{$\mathbf{0.6296}$} \\
            DIDFuse~\cite{ZhaoDIDFuse2020}    &$0.7381$	  &$0.0292$	  &$3.5809$	  &$0.7923$    &$1.1566$	&$0.6575$	&$0.5813$	   &$0.5651$ \\
            AUIF~\cite{zhao2021efficient}   &$0.7324$	&$0.0286$	&\textcolor{cyan}{$\mathbf{3.5821}$}	&$0.7869$	&$1.1436$	&$0.6415$	&$0.5566$	&$0.5700$\\
            SwinFuse~\cite{SwinFuse2022}  &$0.7561$ 	   &\textcolor{cyan}{$\mathbf{0.0361}$}	 &$3.4939$	   &$0.8157$	  &$1.0825$    &\textcolor{cyan}{$\mathbf{0.8891}$}   &$0.5435$   &$0.5860$\\
            YDTR~\cite{Tang_2022_YDTR}   &$0.7350$	 &$0.0332$	  &\textcolor{blue}{$\mathbf{3.5823}$}	&$0.8055$	 &\textcolor{blue}{$\mathbf{1.1845}$}    &$0.7159$   &\textcolor{blue}{$\mathbf{1.0427}$}	   &$0.5840$\\
            TarDAL~\cite{liu2022target}  &\textcolor{cyan}{$\mathbf{0.7608}$}	 &$0.0339$	&$3.5768$	&\textcolor{cyan}{$\mathbf{0.8195}$}	&$1.1492$	&$0.7683$	&\textcolor{cyan}{$\mathbf{1.0278}$}	 &$0.5277$\\
            \bf{MoE-Fusion}   &\textcolor{red}{$\mathbf{0.7688}$}	&\textcolor{red}{$\mathbf{0.0377}$}	  &\textcolor{red}{$\mathbf{3.5884}$}	 &\textcolor{red}{$\mathbf{0.8844}$}	&\textcolor{red}{$\mathbf{1.2664}$}	  &\textcolor{red}{$\mathbf{0.9166}$}	   &$0.8671$	&\textcolor{red}{$\mathbf{0.6964}$}\\
            \midrule
            \multicolumn{9}{c}{\textbf{FLIR Dataset}~\cite{zhang2020multispectral}}    \\
            &EN        &SF         &SD         &MI        &VIF       &AG         &SCD       &$Q_{abf}$       \\ 
            \midrule
            DenseFuse~\cite{li2018densefuse}   &$1.0133$	  &$0.0379$	  &$4.3260$	  &$0.9867$	  &$1.1369$	  &$0.8175$	  &$0.3184$	  &$0.5382$\\
            RFN-Nest~\cite{li2021rfn}      &$1.0472$	  &$0.0333$	  &$4.3269$	  &$0.9998$	  &$1.0739$	  &$0.7711$	  &$0.6335$	  &$0.5086$\\
            IFCNN~\cite{zhang2020ifcnn}    &$1.0570$	  &\textcolor{blue}{$\mathbf{0.0455}$}	  &\textcolor{cyan}{$\mathbf{4.3319}$}	  &$1.0143$	  &$1.1680$	  &$1.1543$	  &\textcolor{cyan}{$\mathbf{0.6894}$}	  &\textcolor{blue}{$\mathbf{0.6434}$}\\
            PIAFusion~\cite{Tang2022PIAFusion}  &\textcolor{cyan}{$\mathbf{1.0615}$}	  &$0.0430$	  &$4.3311$	  &\textcolor{blue}{$\mathbf{1.0197}$}	  &\textcolor{blue}{$\mathbf{1.1706}$}	  &\textcolor{red}{$\mathbf{1.1900}$}	  &$0.6867$	  &\textcolor{cyan}{$\mathbf{0.6214}$}\\
            DIDFuse~\cite{ZhaoDIDFuse2020}    &$1.0439$	  &\textcolor{red}{$\mathbf{0.0505}$}	  &$4.3315$	  &$0.9947$	  &$1.1637$	  &$1.1286$	  &$0.6428$	  &$0.5602$\\
            AUIF~\cite{zhao2021efficient}   &$1.0579$	  &$0.0301$	  &$4.1833$	  &$1.0060$	  &$0.7889$	  &$0.8856$	  &$0.4076$	  &$0.3687$\\
            SwinFuse~\cite{SwinFuse2022}  &\textcolor{blue}{$\mathbf{1.0617}$}	  &$0.0436$	  &$4.3250$	  &$1.0141$	  &$1.1453$	  &\textcolor{blue}{$\mathbf{1.1897}$}	  &\textcolor{blue}{$\mathbf{0.6930}$}	  &$0.5973$\\
            YDTR~\cite{Tang_2022_YDTR}   &$1.0301$	  &$0.0407$	  &\textcolor{blue}{$\mathbf{4.3321}$}	  &$1.0065$	  &\textcolor{cyan}{$\mathbf{1.1698}$}	  &$0.9667$	  &$0.6484$	  &$0.5746$\\
            TarDAL~\cite{liu2022target}  &$1.0605$	  &$0.0381$	  &$4.3305$	  &\textcolor{cyan}{$\mathbf{1.0181}$}	  &$1.0580$	  &$0.9666$	  &$0.5872$	  &$0.5428$\\
            \bf{MoE-Fusion}   &\textcolor{red}{$\mathbf{1.0621}$}	 &\textcolor{cyan}{$\mathbf{0.0444}$}	 &\textcolor{red}{$\mathbf{4.3322}$}  &\textcolor{red}{$\mathbf{1.0203}$}	 &\textcolor{red}{$\mathbf{1.1710}$}	  &\textcolor{cyan}{$\mathbf{1.1548}$}	 &\textcolor{red}{$\mathbf{0.6980}$}	 &\textcolor{red}{$\mathbf{0.6460}$}\\
            \midrule
            \multicolumn{9}{c}{\textbf{LLVIP Dataset}~\cite{jia2021llvip}} \\
            &EN        &SF         &SD         &MI        &VIF       &AG         &SCD       &$Q_{abf}$       \\ 
            \midrule
            DenseFuse~\cite{li2018densefuse}   &$0.5527$	&$0.0172$	&$2.9714$	&$0.6360$	&$1.0694$	&$0.3460$	&$0.5545$	&$0.4465$\\
            RFN-Nest~\cite{li2021rfn}    &$0.5613$	  &$0.0175$	  &$2.9028$	  &$0.6275$	 &$1.0402$	 &$0.3286$	  &$0.7709$	  &$0.3657$\\
            IFCNN~\cite{zhang2020ifcnn}      &$0.5711$	  &$	0.0252$	&$	3.0012$	&$	0.6471$	&$	1.1080$	&$	0.5475$	&$	0.8209$	&\textcolor{cyan}{$\mathbf{0.6708}$}\\
            PIAFusion~\cite{Tang2022PIAFusion}   &\textcolor{red}{$\mathbf{0.5850}$}	 &\textcolor{cyan}{$\mathbf{0.0281}$}	&$2.9529$	&\textcolor{red}{$\mathbf{0.6737}$}	  &\textcolor{blue}{$\mathbf{1.1472}$}	&\textcolor{cyan}{$\mathbf{0.5818}$}	&\textcolor{cyan}{$\mathbf{1.1314}$}	&\textcolor{blue}{$\mathbf{0.6791}$}\\
            DIDFuse~\cite{ZhaoDIDFuse2020}    &$0.5315$	 &$0.0205$	&$2.5029$	&$0.6175$	&$0.8252$	&$0.4030$	&$0.1454$	&$0.3497$\\
            AUIF~\cite{zhao2021efficient}      &$0.5364$	&$0.0199$	&$2.6871$	&$0.6210$	&$0.9058$	&$0.3824$	&$0.1922$	&$0.3306$\\
            SwinFuse~\cite{SwinFuse2022}   &$0.5072$	&$0.0196$	&$2.4646$	&$0.5753$	&$0.8285$	&$0.3697$	&$0.0557$	&$0.2830$\\
            YDTR~\cite{Tang_2022_YDTR}    &$0.5544$	   &$0.0213$	&\textcolor{cyan}{$\mathbf{2.9899}$}	&$0.6470$	 &$1.0960$	  &$0.4096$	   &$0.8406$	&$0.5286$\\
            TarDAL~\cite{liu2022target}  &\textcolor{blue}{$\mathbf{0.5818}$}	&\textcolor{blue}{$\mathbf{0.0283}$}	&\textcolor{blue}{$\mathbf{2.9914}$}	&\textcolor{blue}{$\mathbf{0.6713}$}	&\textcolor{cyan}{$\mathbf{1.1403}$}	&\textcolor{blue}{$\mathbf{0.5820}$}	&\textcolor{blue}{$\mathbf{1.1316}$}	&$0.5957$\\
            \bf{MoE-Fusion}    &\textcolor{cyan}{$\mathbf{0.5725}$}	  &\textcolor{red}{$\mathbf{0.0285}$}	&\textcolor{red}{$\mathbf{3.0074}$}	  &\textcolor{cyan}{$\mathbf{0.6522}$}	&\textcolor{red}{$\mathbf{1.1475}$}	   &\textcolor{red}{$\mathbf{0.5824}$} 	&\textcolor{red}{$\mathbf{1.1317}$}	  &\textcolor{red}{$\mathbf{0.6821}$}\\
            \bottomrule
        \end{tabular}}%
        \vspace{2.0cm}
\end{table*}

\begin{table*}[t]
    \centering
    \caption{Local Quantitative comparison of our MoE-Fusion with $9$ state-of-the-art methods for {\bf the background regions of fused images}. Bold \textcolor{red}{red} indicates the best, Bold \textcolor{blue}{blue} indicates the second best, and Bold \textcolor{cyan}{cyan} indicates the third best.}
    \vspace{0.2cm}
    \label{tab:BG_SOTA}%
    \resizebox{\linewidth}{!}{
        \begin{tabular}{ccccccccc}
            \toprule
            \multicolumn{9}{c}{\textbf{M$^{3}$FD Dataset}~\cite{liu2022target}}        \\
            &EN        &SF         &SD         &MI        &VIF       &AG         &SCD       &$Q_{abf}$       \\ 
            \midrule
            DenseFuse~\cite{li2018densefuse}   &$6.2466$	&$0.0413$	 &$9.3869$	   &$3.4312$	  &$0.8207$	  &$2.9234$	  &$1.2845$	   &$0.4216$\\
            RFN-Nest~\cite{li2021rfn}      &$6.7317$   &$0.0396$	&$9.7022$	  &$3.4181$	  &$0.9304$	   &$3.0006$	&$1.4032$	 &$0.4115$\\
            IFCNN~\cite{zhang2020ifcnn}    &$6.4615$	  &$0.0611$	 &$10.1760$	  &$3.4543$	  &$0.9117$	&$4.7119$	&$1.3914$    &\textcolor{blue}{$\mathbf{0.5948}$}\\
            PIAFusion~\cite{Tang2022PIAFusion}  &$6.6101$	  &\textcolor{blue}{$\mathbf{0.0697}$}	   &\textcolor{blue}{$\mathbf{10.6753}$}   &\textcolor{blue}{$\mathbf{4.2777}$}  &\textcolor{cyan}{$\mathbf{0.9967}$}   &\textcolor{blue}{$\mathbf{5.1920}$}   &$1.1991$    &\textcolor{cyan}{$\mathbf{0.5724}$}\\
            DIDFuse~\cite{ZhaoDIDFuse2020}    &$6.4335$	  &$0.0463$	  &$9.9337$	  &$3.4657$    &$0.8963$	&$3.3748$	 &$1.3576$	   &$0.4667$ \\
            AUIF~\cite{zhao2021efficient}   &$6.3530$	&$0.0439$	&$9.5516$	&$3.4630$	&$0.8432$	&$3.1355$	&$1.3180$	 &$0.4426$\\
            SwinFuse~\cite{SwinFuse2022}  &\textcolor{red}{$\mathbf{7.0682}$} 	   &\textcolor{cyan}{$\mathbf{0.0674}$}	   &$9.7274$	   &$3.6591$	  &\textcolor{blue}{$\mathbf{1.0340}$}    &\textcolor{cyan}{$\mathbf{5.1878}$}   &\textcolor{red}{$\mathbf{1.4323}$}   &$0.5267$\\
            YDTR~\cite{Tang_2022_YDTR}   &$6.3607$	 &$0.0541$	  &\textcolor{cyan}{$\mathbf{10.2420}$}	  &$3.6616$	  &$0.8902$    &$3.7014$   &\textcolor{cyan}{$\mathbf{1.4174}$}	   &$0.5120$\\
            TarDAL~\cite{liu2022target}  &\textcolor{blue}{$\mathbf{6.9323}$}	  &$0.0551$	  &$10.1222$ &\textcolor{cyan}{$\mathbf{3.7133}$}	&$0.9856$	 &$3.9125$	  &$1.3664$	   &$0.4201$\\
            \bf{MoE-Fusion}   &\textcolor{cyan}{$\mathbf{6.7860}$}	&\textcolor{red}{$\mathbf{0.0710}$}	  &\textcolor{red}{$\mathbf{10.6973}$}	&\textcolor{red}{$\mathbf{4.5515}$}	&\textcolor{red}{$\mathbf{1.1308}$}	  &\textcolor{red}{$\mathbf{5.2203}$}	   &\textcolor{blue}{$\mathbf{1.4284}$}	&\textcolor{red}{$\mathbf{0.6721}$}\\
            \midrule
            \multicolumn{9}{c}{\textbf{FLIR Dataset}~\cite{zhang2020multispectral}}    \\
            &EN        &SF         &SD         &MI        &VIF       &AG         &SCD       &$Q_{abf}$       \\ 
            \midrule
            DenseFuse~\cite{li2018densefuse}   &$6.7076$	  &$0.0455$	  &$10.9301$	  &$3.5629$	  &$0.7998$	  &$2.9991$	  &$1.0409$	  &$0.3475$\\
            RFN-Nest~\cite{li2021rfn}      &\textcolor{cyan}{$\mathbf{7.1380}$}	  &$0.0395$	  &$10.9758$	  &$3.5958$	  &$0.8616$	  &$2.6326$	  &$1.1256$	  &$0.3046$\\
            IFCNN~\cite{zhang2020ifcnn}    &$6.8659$	  &$0.0674$	  &$10.8812$	  &$3.5080$	  &$0.9094$	&\textcolor{blue}{$\mathbf{5.3579}$}	  &$1.1797$	   &\textcolor{red}{$\mathbf{0.5624}$}\\
            PIAFusion~\cite{Tang2022PIAFusion}  &$6.6917$	  &$0.0648$	  &\textcolor{blue}{$\mathbf{11.8801}$}	  &$3.5812$	&\textcolor{blue}{$\mathbf{0.9594}$}	&$4.8385$	  &$1.0870$	   &\textcolor{cyan}{$\mathbf{0.4676}$}\\
            DIDFuse~\cite{ZhaoDIDFuse2020}    &$6.9978$	  &\textcolor{blue}{$\mathbf{0.0680}$}	 &\textcolor{red}{$\mathbf{11.8999}$}	  &$3.1025$	  &$0.8305$	  &\textcolor{cyan}{$\mathbf{5.3311}$}	&\textcolor{blue}{$\mathbf{1.4191}$}	&$0.3856$  \\
            AUIF~\cite{zhao2021efficient}   &$7.0057$	  &$0.0468$	  &$10.1748$	  &$3.5235$	  &$0.7845$	  &$3.9412$	  &$0.6594$	  &$0.3244$\\
            SwinFuse~\cite{SwinFuse2022}  &$7.1181$ 	&\textcolor{cyan}{$\mathbf{0.0679}$}	  &$10.9948$	  &$3.5513$	&\textcolor{cyan}{$\mathbf{0.9579}$}	  &$5.2656$	    &\textcolor{red}{$\mathbf{1.5214}$}	 	  &$0.4090$\\
            YDTR~\cite{Tang_2022_YDTR}   &$6.6353$	  &$0.0534$	  &$11.0152$    &\textcolor{cyan}{$\mathbf{3.6865}$}   &$0.8334$	  &$3.2230$	  &$1.2674$	  &$0.3743$\\
            TarDAL~\cite{liu2022target}  &\textcolor{blue}{$\mathbf{7.2068}$}	  &$0.0640$	  &$10.9833$	&\textcolor{blue}{$\mathbf{3.7278}$}	  &$0.9042$	  &$4.8738$	  &$0.8469$	  &$0.4323$\\
            \bf{MoE-Fusion}      &\textcolor{red}{$\mathbf{7.2187}$}	&\textcolor{red}{$\mathbf{0.0681}$}	 &\textcolor{cyan}{$\mathbf{11.0548}$}  &\textcolor{red}{$\mathbf{3.7368}$}	&\textcolor{red}{$\mathbf{0.9667}$}	  &\textcolor{red}{$\mathbf{5.3910}$}	 &\textcolor{cyan}{$\mathbf{1.4059}$}	&\textcolor{blue}{$\mathbf{0.5246}$}\\
            \midrule
            \multicolumn{9}{c}{\textbf{LLVIP Dataset}~\cite{jia2021llvip}} \\
            &EN        &SF         &SD         &MI        &VIF       &AG         &SCD       &$Q_{abf}$       \\ 
            \midrule
            DenseFuse~\cite{li2018densefuse}   &$6.7613$	&$0.0434$	&$9.5516$	&$3.0287$	&$0.7521$	&$3.1273$	&$1.1330$	&$0.3181$\\
            RFN-Nest~\cite{li2021rfn}    &$7.0563$	&$0.0330$	&$9.8548$	&$2.8719$	&$0.7955$	&$2.6958$	&\textcolor{cyan}{$\mathbf{1.4070}$}	&$0.2460$\\
            IFCNN~\cite{zhang2020ifcnn}      &$7.1119$	 &\textcolor{cyan}{$\mathbf{0.0680}$}	 &\textcolor{blue}{$\mathbf{9.9317}$}	&$3.2737$    &\textcolor{cyan}{$\mathbf{0.8395}$}	 &\textcolor{cyan}{$\mathbf{5.1106}$}    &$1.3752$	  &\textcolor{red}{$\mathbf{0.5958}$}\\
            PIAFusion~\cite{Tang2022PIAFusion} &\textcolor{red}{$\mathbf{7.3004}$}	  &\textcolor{blue}{$\mathbf{0.0784}$}	  &$9.8792$	  &\textcolor{blue}{$\mathbf{3.6756}$}	  &\textcolor{blue}{$\mathbf{0.9380}$}	  &\textcolor{blue}{$\mathbf{5.7825}$}	  &\textcolor{blue}{$\mathbf{1.5195}$}	  &\textcolor{cyan}{$\mathbf{0.5772}$}\\
            DIDFuse~\cite{ZhaoDIDFuse2020}    &$5.9448$	  &$0.0527$	  &$7.6424$	  &$2.8137$	   &$0.5465$	 &$3.1695$	  &$1.0767$	  &$0.2340$\\
            AUIF~\cite{zhao2021efficient}      &$6.1165$	  &$0.0617$	  &$7.7777$	  &$2.7252$	  &$0.6080$	  &$3.6030$	  &$1.1095$	  &$0.2725$\\
            SwinFuse~\cite{SwinFuse2022}   &$5.8519$	 &$0.0589$	  &$7.5822$	  &$2.4033$	  &$0.6421$	  &$3.4821$	  &$1.0832$	  &$0.2630$\\
            YDTR~\cite{Tang_2022_YDTR}    &$6.6232$	  &$0.0484$	  &$9.1427$	  &$3.2533$	  &$0.6902$	  &$3.0566$	  &$1.0642$	  &$0.2918$\\
            TarDAL~\cite{liu2022target}  &\textcolor{blue}{$\mathbf{7.2692}$}	  &$0.0679$	  &\textcolor{cyan}{$\mathbf{9.8993}$}	  &\textcolor{red}{$\mathbf{3.7244}$}	  &$0.8328$	   &$4.4663$	  &$1.3791$	   &$0.4398$\\
            \bf{MoE-Fusion}             &\textcolor{cyan}{$\mathbf{7.2048}$}	  &\textcolor{red}{$\mathbf{0.0853}$}	&\textcolor{red}{$\mathbf{9.9427}$}	  &\textcolor{cyan}{$\mathbf{3.3231}$}	&\textcolor{red}{$\mathbf{0.9678}$}	   &\textcolor{red}{$\mathbf{5.8198}$} 	&\textcolor{red}{$\mathbf{1.7371}$}	  &\textcolor{blue}{$\mathbf{0.5899}$}\\
            \bottomrule
        \end{tabular}}%
        \vspace{2.0cm}
\end{table*}

\subsection{Local Comparison on the LLVIP Dataset}
According to Table~\ref{tab:FG_SOTA}, it can be seen that we achieved significant advantages on $6$ metrics.
Specifically, the highest SF, AG, and SD illustrate that our fusion results can have the richest texture detail and the highest contrast information in the local foreground region.
The superior performance on SCD and $Q_{abf}$ also indicates that our fusion results on the local foreground region can effectively learn complementary and edge information from multi-modal images. Moreover, the highest VIF also demonstrates that our fusion results are more favourable to human observation on the local foreground region. These results illustrate that the proposed method can effectively preserve valuable information in the local foreground region of the multi-modal images, which benefits from the specialized learning of local information.

As seen in Table~\ref{tab:BG_SOTA}, we achieved the best on $5$ metrics and the second best on $1$ metric.
The highest SF, AG, SD, and SCD indicate that our method preserves the richest texture details and valuable multi-modal complementary information on the local background region of fused images. The highest VIF also indicates that our fused image is closer to human vision on the local background region. In addition, the second best on $Q_{abf}$ also illustrates that our method is competitive in preserving multi-modal edge information. These results demonstrate that our method can better motivate fused images to preserve local background detail information of multi-modal images through sample-adaptive dynamic learning.

\section{Encoder Architecture}
\begin{figure}[!t]
	\centering
	\includegraphics[width=1.0\columnwidth]{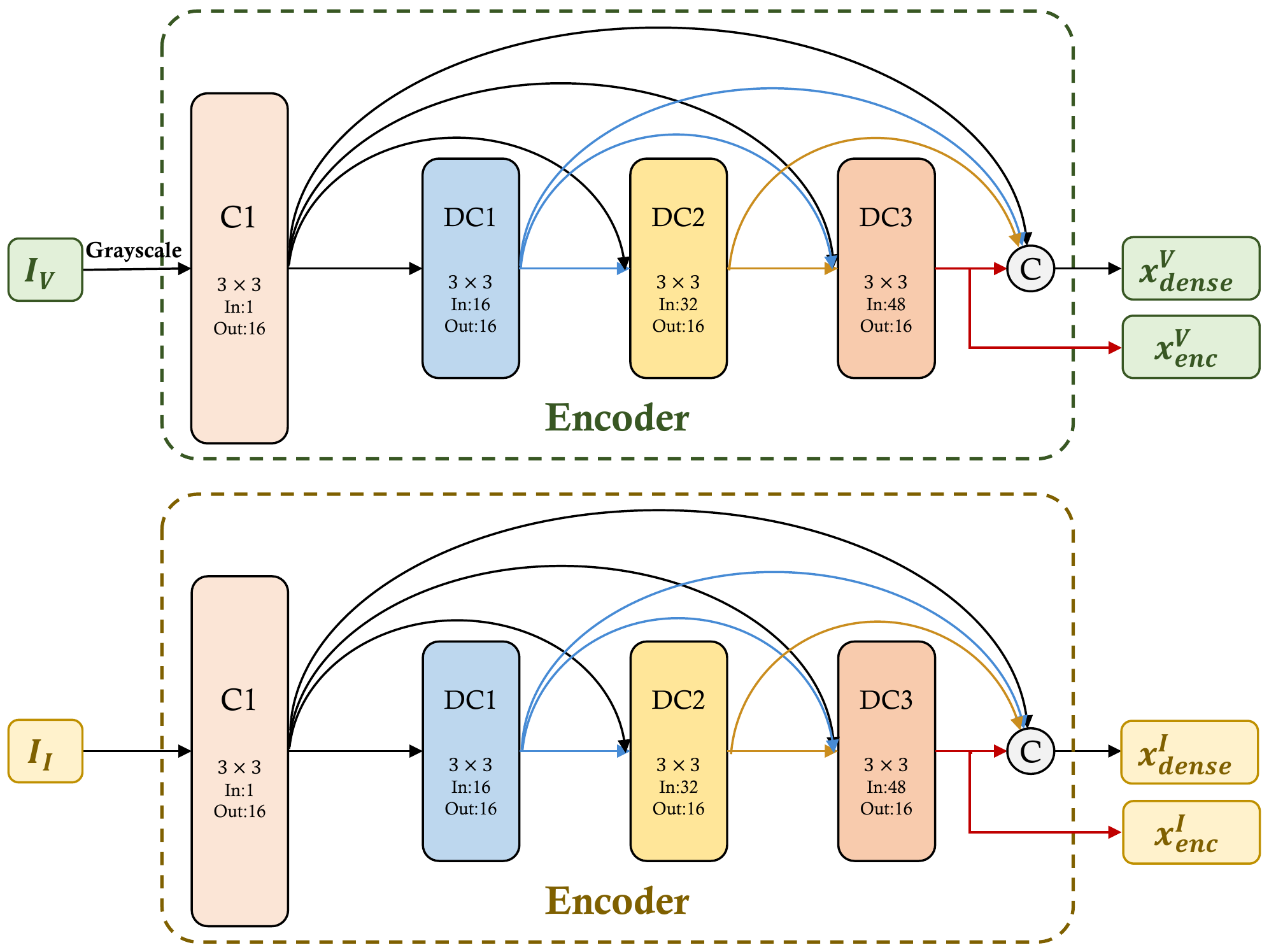}
	\caption{The architecture of the two encoders in the proposed MoE-Fusion.}
	\label{fig:Encoder}
        \vspace{-0.3cm}
\end{figure}

The detailed architecture of the two encoders in the proposed MoE-Fusion is shown in Fig.~\ref{fig:Encoder}. By convention, we grayscale the $3$-channel visible image $I_\mathcal{V}$ to obtain a single-channel image, and then send it and the infrared image $I_\mathcal{I}$ to the two encoders ($Enc_\mathcal{V}$ and $Enc_\mathcal{I}$) separately for feature extraction. These two encoders have the same structure, but the parameters are not shared. Each encoder contains one convolutional layer $C1$ and three densely connected convolutional layers ($DC1$, $DC2$, $DC3$). Referring to~\cite{huang2017densely}, the output of each convolutional layer is fed into each subsequent convolutional layer in the densely connected mechanism, which facilitates the preservation of deep features as much as possible. Finally, we obtain the feature maps of $DC3$ layer ($x_{enc}^\mathcal{I}$ and $x_{enc}^\mathcal{V}$) and dense feature maps ($x_{dense}^\mathcal{I}$ and $x_{dense}^\mathcal{V}$) from the two encoders, respectively.

\section{Fusion Loss}
In MoE-Fusion, the fusion loss $\mathcal{L}_{fusion}$ contains the pixel loss $\mathcal{L}_{pixel}$, gradient loss $\mathcal{L}_{grad}$, and load loss $\mathcal{L}_{load}$. We provide more details on their formalization in this supplementary material.

\subsection{Pixel Loss}
We perform the different operations on the foreground and background of the multi-modal images to facilitate the fusion network to learn more valuable pixel intensity information. $\mathcal{L}_{pixel}$ is defined as,
\begin{align}
\mathcal{L}_{pixel}= \mathcal{L}_{pixel}^{fg}+ \mathcal{L}_{pixel}^{bg},
\end{align}
where $\mathcal{L}_{pixel}^{fg}$ represents the pixel loss on the foreground regions, and $\mathcal{L}_{pixel}^{bg}$ represents the pixel loss on the background regions. Their formalizations are as follows,
\begin{align}
\mathcal{L}_{pixel}^{fg} =\frac{1}{HW} \left \| I_{m} \circ \left ( I_\mathcal{F} - max\left ( I_\mathcal{V} , I_\mathcal{I} \right ) \right ) \right \| _{1},
\end{align}
\begin{align}
\mathcal{L}_{pixel}^{bg} =\frac{1}{HW} \left \| \left ( 1- I_{m} \right ) \circ \left ( I_\mathcal{F} - mean\left ( I_\mathcal{V} , I_\mathcal{I} \right ) \right ) \right \| _{1},
\end{align}
where $H$ and $W$ represent the height and width of the image, respectively, $I_{m}$ is the foreground mask generated according to the ground-truth bounding boxes of the auxiliary detection network.
\begin{math}
  \left \| \cdot  \right \| _{1}
\end{math} stands for the $l_{1}$-norm, the operator $\circ$ denotes the element-wise multiplication, max($\cdot$) denotes the element-wise maximum operation, and mean($\cdot$) denotes the element-wise average operation. 

\subsection{Gradient Loss}
We expect the fused image to preserve the richest texture details of the images from both modalities. So the gradient loss $\mathcal{L}_{grad}$ is formulated as,
\begin{align}
\mathcal{L}_{grad}=\frac{1}{HW} \left \| \left | \nabla I_\mathcal{F} \right | - max\left ( \left | \nabla I_\mathcal{V} \right |,\left | \nabla I_\mathcal{I} \right | \right ) \right \| _{1},
\end{align}
where $\nabla$ denotes the Sobel gradient operator, which measures the texture detail information of an image. $\left | \cdot \right |$ stands for the absolute operation.

\subsection{Load Loss}
In the Mixture-of-Experts (MoE), the load loss is mainly used to encourage experts to receive roughly equal numbers of training examples. The proposed MoE-Fusion contains MoLE and MoGE. Therefore, the load loss $\mathcal{L}_{load}$ in our work consists of two parts, namely $\mathcal{L}_{load}^{local}$ and $\mathcal{L}_{load}^{global}$. The $\mathcal{L}_{load}$ can be calculated as: 
\begin{align}
  \mathcal{L}_{load}= \mathcal{L}_{load}^{local} + \mathcal{L}_{load}^{global}.
  \label{eq:21}
\end{align}

Following~\cite{Shazeer2017OutrageouslyLN}, the $\mathcal{L}_{load}^{local}$ and $\mathcal{L}_{load}^{global}$ are defined as the square of the coefficient of variation of the load vector. We also follow~\cite{Shazeer2017OutrageouslyLN} to initialize the weight matrix of the gate network in each MoE to zero, which keeps the expert load of MoE in an approximately equal state during the initial phase.

\section{Color Space Conversion of Fused Images}
By convention, the output of a fusion network is a single-channel image. We can better present the fusion results by supplementing color information to the fused image through color space conversion. The color information is mainly preserved in the visible images. We first transfer the visible image from the RGB color space to the YCbCr color space and extract the Cb and Cr channels of the visible image, which contain the color information. Then we concatenate the single-channel fused image with the Cb and Cr channels of the visible image to obtain the $3$-channel fused image. Finally, we convert the $3$-channel fused image back to the RGB color space to obtain the color fusion result.

\section{Limitation}
In this work, the total number of experts $N$, as well as the number of sparsely activated experts $K$ in MoLE and MoGE are selected empirically. The potential of our model may not be fully exploited due to the limitation of the data scale and computational resources. Considering the dynamic scenarios in reality, the empirical selection may restrict the fusion performance in broader scenarios. Therefore, how to adaptively select the most suitable numbers of experts $N$ and sparse activated experts $K$ according to the scenario is still a challenging problem. We will study to overcome it in future work.

{\small
\bibliographystyle{ieee_fullname}
\bibliography{egbib}
}

\end{document}